%% file: nips_cfl.tex
\documentclass[journal]{IEEEtran}

\usepackage[utf8]{inputenc} 
\usepackage[T1]{fontenc}    
\usepackage{hyperref}       
\usepackage{url}            
\usepackage{booktabs}       
\usepackage{amsfonts}       
\usepackage{amsmath}
\usepackage{amsthm}
\usepackage{ amssymb }
\usepackage{nicefrac}       
\usepackage{microtype}      
\usepackage{graphicx}
\usepackage{float}
\usepackage{comment}
\usepackage[ruled, linesnumbered]{algorithm2e}
\usepackage{tabularx}
\usepackage{multirow}
\usepackage{bbm}
\usepackage{makecell}
\usepackage{wrapfig}
\usepackage{booktabs}
\usepackage{array}
\usepackage{algorithmic}
\usepackage{graphicx}
\usepackage{textcomp}
\usepackage{xcolor}
\usepackage{subcaption}

\usepackage{tikz}

\DeclareMathOperator*{\argmax}{arg\,max}

\theoremstyle{definition}

\title{Communication-Efficient Federated Distillation}

\author{
  Felix Sattler$^*$, 
  Arturo Marban, Roman Rischke, and
   Wojciech Samek$^*$,~\IEEEmembership{Member,~IEEE}
\thanks{\textsuperscript{*}~Corresponding authors: F.~Sattler and W.~Samek.}%
\thanks{This work was supported by the Federal Ministry of Education and Research (BMBF) through the BIFOLD - Berlin Institute for the Foundations of Learning and Data (ref.~01IS18025A and ref~01IS18037I).}
\thanks{F.~Sattler, A.~Marban, R.~Rischke, and W.~Samek is with the Fraunhofer Heinrich Hertz Institute, 10587 Berlin, Germany (e-mail: 
\href{mailto:felix.sattler@hhi.fraunhofer.de}{felix.sattler@hhi.fraunhofer.de}, 
\href{mailto:wojciech.samek@hhi.fraunhofer.de}{wojciech.samek@hhi.fraunhofer.de}).}}

\begin{document}

\parskip 0pt

\maketitle

\begin{abstract}
Communication constraints are one of the major challenges preventing the wide-spread adoption of Federated Learning systems. Recently, Federated Distillation (FD), a new algorithmic paradigm for Federated Learning with fundamentally different communication properties, emerged. FD methods leverage ensemble distillation techniques and exchange model outputs, presented as soft labels on an unlabeled public data set, between the central server and the participating clients. While for conventional Federated Learning algorithms, like Federated Averaging (FA), communication scales with the size of the jointly trained model, in FD communication scales with the distillation data set size, resulting in advantageous communication properties, especially when large models are trained. In this work, we investigate FD from the perspective of communication efficiency by analyzing the effects of active distillation-data curation, soft-label quantization and delta-coding techniques. Based on the insights gathered from this analysis, we present Compressed Federated Distillation (CFD), an efficient Federated Distillation method. 
Extensive experiments on Federated image classification and language modeling problems demonstrate that our method can reduce the amount of communication necessary to achieve fixed performance targets by more than two orders of magnitude, when compared to FD and by more than four orders of magnitude when compared with FA.


\end{abstract}

\section{Introduction}
\label{sec:intro}

As many cases of data leakage and misuse in recent times have demonstrated, the centralized processing of personal user data in the "cloud" (e.g.\ for training deep learning models) is associated with a high privacy risk for the data donors. To address this issue, recently a novel distributed training paradigm called Federated Learning (FL) emerged.

FL \cite{mcmahan2017communication}\cite{kairouz2019advances}\cite{li2020federated} 
allows multiple entities to jointly train a machine learning model on their combined data, without any of the participants having to reveal their potentially privacy sensitive data to a centralized server. Federated Learning achieves this, by processing the data on the local devices and only communicating sanitized or encrypted information about the underlying patterns to other devices and the server.

Besides improving privacy, FL comes with many other benefits such as improved security \cite{ma2020safeguarding}, autonomy \cite{niknam2020federated} and efficiency \cite{sattler2020trends} due to its distributed nature and on-device processing. 

FL is typically performed between mobile and internet of things (IoT) devices, which are often severely hardware constrained, geographically scattered and have only access to limited and costly communication channels like metered mobile networks. 
To harness the ever growing amounts of privacy sensitive data collected by these devices, there is thus great need for efficient and scalable FL solutions.

One of the most challenging obstacles in Federated Learning, is the communication bottleneck induced by frequently exchanging training information between the participating clients over limited bandwidth channels. For instance, the communication of local gradients, which are the basic unit of information for gradient descent based distributed training methods like distributed SGD, requires $\mathcal{O}(|\theta|)$ bits of information, where $|\theta|$ is the model size. Over the course of multiple thousands of training rounds the communication overhead can grow to hundreds of Gigabytes for modern large-scale neural-network models with millions of parameters. Consequently, if communication bandwidth is limited or communication is costly, federated learning can become unproductive or even completely unfeasible.

To address this issue, different algorithmic approaches have been proposed under the umbrella of efficient Federated Learning. In this work, we closely examine the recently proposed Federated Distillation method \cite{itahara2020distill} with respect to it's communication properties and propose a set of improvements, which reduce communication in both the upstream and the downstream, without negatively affecting the training performance. More concretely, we make the following contributions:
\begin{itemize}
    \item We conduct a qualitative and quantitative comparison between the communication properties of two popular algorithmic frameworks for Federated Learning, namely Federated Averaging \cite{mcmahan2017communication} and Federated Distillation.
    \item  We perform a thorough analysis of the communication properties of Federated Distillation at different levels of data heterogeneity by investigating the effects of distillation data set size as well as active data selection strategies on the training performance.
    \item We develop a novel quantization mechanism and delta coding method to compress the soft-labels exchanged in Federated Distillation before communication.
    \item We address the issue of compressing downstream communication via a novel dual distillation technique.
    \item Finally, we perform extensive experiments on large-scale convolutional neural networks and transformer models, which demonstrate that our compression method can reduce communication by more than $\times$100 as compared to Federated Distillation and more than $\times$10000 as compared to Federated Averaging.
\end{itemize}

The remainder of this manuscript is organized as follows: In section \ref{sec2} we describe the two major algorithmic frameworks in Federated Learning, namely Federated Averaging and Federated Distillation, compare them w.r.t. their communication properties and review existing techniques for communication reduction in both frameworks. In section \ref{sec3} we thoroughly investigate ways to reduce the communication in Federated Distillation by systematically addressing all components that contribute to the total communication load. In section \ref{section:compressed_fed_distillation} we condense the gathered insights and propose Compressed Federated Distillation (CFD), a novel communication-efficient Federated Distillation scheme. Finally, in section \ref{sec:experiments} we compare the communication properties of CFD with those of regular Federated Distillation and Federated Averaging on a variety of Federated Learning benchmarks featuring large-scale convolutional and transformer neural networks, before concluding in section \ref{sec7}.

\section{Algorithmic Frameworks for Federated Learning}
\label{sec2}
\begin{figure*}
    \centering
    \includegraphics[width=\textwidth]{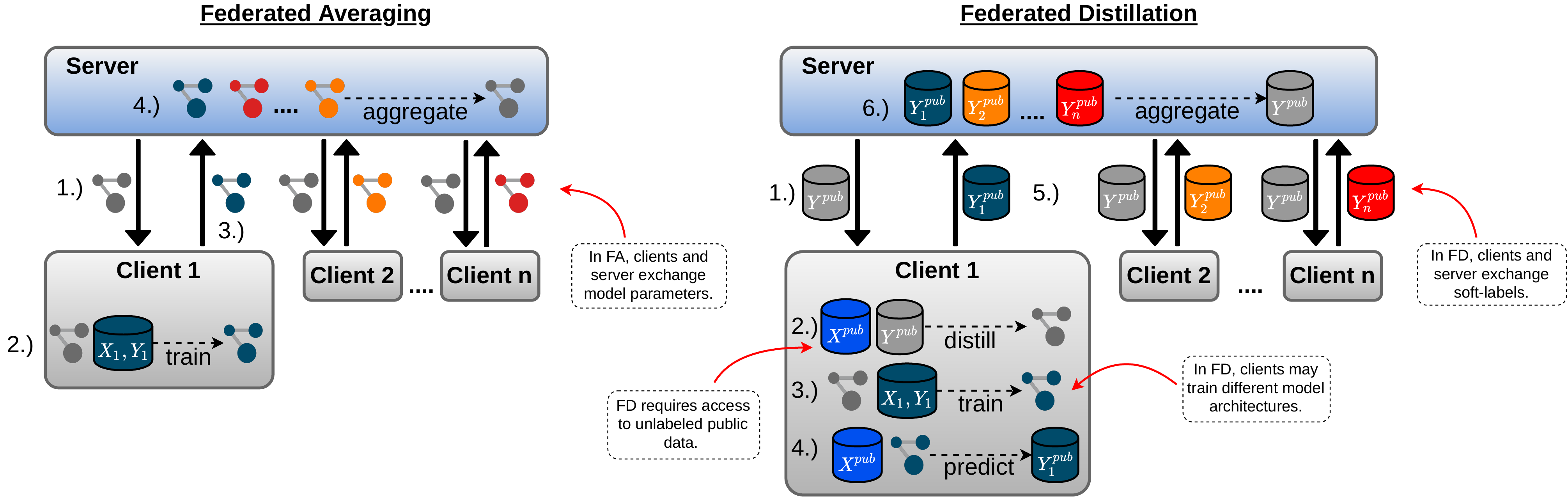} 
    \caption{The flow of data and computations in Federated Averaging and Federated Distillation. In Federated Averaging the model parameters $\theta$ are used to transfer the training information between clients and the server. In Federated Distillation, soft-label predictions $Y^{pub}$ on a common public data set $X^{pub}$ are used to convey the same information. }
    \label{fig:overview}
\end{figure*} 
To solve federated learning problems, two algorithmic frameworks have been proposed, which drastically differ with respect to their communication properties. Figure \ref{fig:overview} gives an overview of the frameworks and compares them w.r.t. to the flow of computation and communication. 
\subsection{Federated Averaging}
The classical algorithmic approach to Federated Learning problems is Federated Averaging \cite{mcmahan2017communication} (Figure \ref{fig:overview} left). In Federated Averaging the training is conducted in multiple communication rounds following a three step protocol:  
\begin{enumerate}
    \item In the beginning of each round, the central server selects a subset of the client population and broadcasts to them a common model initialization $\theta$.
    \item Starting from the common initialization, the selected clients individually perform iterations of stochastic gradient descent over their local data to improve their local models resulting in an updated model $\theta_i$ on every client.
    \item The updated models are then communicated back to the server, where they are aggregated (e.g.\ by an averaging operation) to create a global model, which is used as initialization point for the next communication round.
\end{enumerate}

Every communication round of Federated Averaging thus involves the upstream and downstream communication of a complete parametrization of the jointly trained model $\theta$ between all participating clients and the server.   
In many practical applications these neural network parametrizations may contain multiple millions to billions of individual parameters. For instance, the widely popular ResNet-50 \cite{he2016deep} contains over 23 million parameters. For natural language processing tasks even larger models are used, with the famous GPT-3 \cite{brown2020language} clocking in at 175 billion parameters. Generally, both theoretical \cite{kidger2020uat, chong2020uat} and empirical \cite{huang2019gpipe} evidence suggests that the performance of neural network models correlates positively with their size.   

For large-scale models like the ones described above, the communication overhead of running the Federated Averaging algorithm can become a prohibitive bottleneck.
Although a wide variety of methods to reduce the communication overhead in Federated Averaging have been proposed, including approaches that reduce the frequency of communication \cite{mcmahan2017communication}, use client sampling \cite{mcmahan2017communication, nishio2019clients}, neural network pruning \cite{lecun1990pruning}, message sparsification \cite{aji2017sparse, sattler2018sparse, sattler2019robust} and other lossy \cite{courbariaux2015compress, li2016compress, konevcny2016lowrank, sattler2018sparse, xu2020compress} and loss-less compression techniques \cite{neumann2020deepcabac, wiedemann2020deepcabac}, the fundamental issue of scaling to larger models persists.

\subsection{Federated Distillation}
\label{sec:FD}
The recently proposed Federated Distillation \cite{jeong2018distill, lin2020ensemble, itahara2020distill} (Figure \ref{fig:overview} right) takes an entirely different approach to communicating the knowledge obtained during the local training. Instead of communicating the parameterization of the locally trained model $\theta_i$ to the server, in Federated Distillation the knowledge is communicated in the form of soft-label predictions on records of a public distillation data set $X^{pub}$ according to
\begin{align}
Y_i = \{f_{\theta_i}(x)|x \in X^{pub}\}.    
\end{align}
Hereby $f_{\theta_i}$ is the (neural network) model parametrized by $\theta_i$.
Prior work \cite{li2020federated} has shown that this public distillation data needs to only roughly follow a similar distribution as the privacy sensitive client data and that generally a wide variety of data sets can be suitable to pose as distillation data. For instance, in many federated computer vision problems, extremely large image corpora like ImageNet \cite{deng2009imagenet} are publicly available. Likewise for natural language processing problems public text corpora like WiKiText \cite{merity2016pointer} can be found. While this public data is typically unfit for training a task-specific model due to missing label information, it can still be useful in Federated Distillation pipelines.    

Different variations of Federated Distillation have been proposed that vary w.r.t.\ their communication properties. To fully appreciate the communication saving benefits of Federated Distillation, it is necessary to avoid communication of model parametrizations at all stages of Federated training. We therefore consider the following version of the Federated Distillation protocol for which each communication round consists of the following five steps:
\begin{enumerate}
    \item At the beginning of every Federated Distillation round, a subset of the client population is selected for participation and synchronizes with the server by downloading aggregated soft-labels $Y^{pub}$ on the public data set.
    \item The participating clients update their local models by performing model distillation using the downloaded soft-label information. All stochasticity in the distillation process is controlled via random seeds to ensure that all clients end up with the same distilled model $\theta$.  
    \item The participating clients improve the distilled model by training on their private local data, resulting in improved models $\theta_i$ on every client.
    \item Using the locally trained model $\theta_i$, the clients compute soft-labels $Y_i^{pub}$ on the public data and send them to the server.
    \item The server aggregates the soft-labels for the next communication round.
    
\end{enumerate}

This protocol is most similar to what has been proposed in \cite{itahara2020distill}.

As demonstrated in recent studies \cite{jeong2018distill, lin2020ensemble, itahara2020distill}, Federated Distillation has several advantages over Federated Averaging: First, as model information is aggregated by means of distillation, Federated Distillation allows the participating clients to train different model architectures. This gives additional flexibility in settings where clients have heterogeneous hardware constraints. Federated Distillation also benefits from increased robustness, as adversarial or malicious clients can not directly influence the parametrization of the jointly trained model (only indirectly via their soft-labels). The most significant advantage, however, arises from the fact that Federated Distillation has a completely different communication profile than Federated Averaging. While the upstream and downstream communication in every round of Federated Averaging scales with the size of the jointly trained neural network as
\begin{align}
    \mathtt{b} \in \mathcal{O}(|\theta|))
\end{align}
in Federated Distillation communication scales with the product of the distillation data set size $|X^{pub}|$ and the number of different classes $\text{dim}(\mathcal{Y})$ as
\begin{align}
    \mathtt{b} \in \mathcal{O}(|X^{pub}|\text{dim}(\mathcal{Y})).
\end{align}
 
This can put FD at an advantage in applications where large neural networks are trained, as is the case for instance in natural language processing and computer vision tasks (among many other application). 

Nevertheless Federated Distillation is still communication intensive, especially for large multi-class tasks where sizable distillation data sets are used.

The aim of this work is thus to further improve the communication efficiency in FD, by exploring a variety of communication reduction techniques. Our efforts will culminate in the development of our Compressed Federated Distillation (CFD) method, a novel compression technique for FD based on soft-label quantization, delta coding and dual distillation.

\section{Related Work}
Albeit their novelty, Federated Distillation techniques have been used in several existing works already. To avoid confusion, in the following we present a comprehensive overview on these existing techniques.
Most relevant for the studied multi-round protocol for diverse models in this paper is the protocol proposed by Itahara et al. \cite{itahara2020distill}, which is based on ideas from Jeong et al. \cite{jeong2018distill} and mostly follows the steps described in section \ref{sec:FD} with the sole exception that client models are required to participate in every round and are not kept synchronized during local distillation by means of random seeds.
The similar protocol by Jeong et al. \cite{jeong2018distill} and Seo et al. \cite{seo2020fd} is instead based on locally accumulated logits per \emph{label}, which are aggregated by the server. Furthermore, instead of exploiting these global logits for refining the local models by direct distillation, they are used for regularizing the local training in the next round. 
Similarly, Bistritz et al. \cite{bistritz2020peer2peer} use distillation on an unlabelled public dataset for regularizing on-device learning in a peer-to-peer network.
Guha et al. \cite{guha2019one-shot} propose a one-shot distillation method for convex models, where the server distills the locally optimized client models in a single round based on an unlabelled data set.

The recently proposed FedMD by Li and Wang \cite{li2019fedmd} and Cronus by Chang et al. \cite{chang2019cronus} also address knowledge distillation in Federated Learning through aggregated logits for a public dataset. In FedMD, the clients train in each round first on the public dataset and then on the private dataset for personalization and communicate afterwards their model output on the public dataset to the server, where the aggregation of the uploaded logits for the next round is performed. For the initial pretraining in FedMD the public dataset is required to be labelled, whereas in the communication rounds after initialization the aggregated logits from the clients serve as soft-labels for the public dataset. In Cronus, however, each client uses the local dataset and the soft-labelled public dataset jointly for local training.

Lin et al.~\cite{lin2020ensemble} apply ensemble distillation on top of Federated Averaging to refine the global server model resulting in fewer communication rounds compared to benchmark Federated Averaging methods. Although leveraging the power of ensemble distillation for robust model fusion and data augmentation, their method, called FedDF, is based on the classical Federated Averaging protocol with all the mentioned consequences w.r.t.\ the communication-efficiency. 

Chen and Chao \cite{chen2020bayesian} introduce FedBE, where the server creates Bayesian model ensembles based on the uploaded client models, instead of directly averaging the client models as in FedAvg, and uses an unlabelled dataset to distill one global student model from the Bayesian teacher models. This global model is transferred back to the clients as initialization for the next round of local training. Although using distillation aiming to improve the global server model, their approach is closer to classical Federated Averaging than to Federated Ensemble Distillation, since all clients have to train the same model architecture and the model parameters are communicated up- and downstream. 

We are not aware of any prior work the aims to directly improve the efficiency of the Federated Distillation process by means of compressing the soft-label information. 
We initiate this study and hope to foster further research in this direction for this new algorithmic paradigm in Federated Learning.

\begin{figure}
    \centering
    \includegraphics[width=0.5\textwidth]{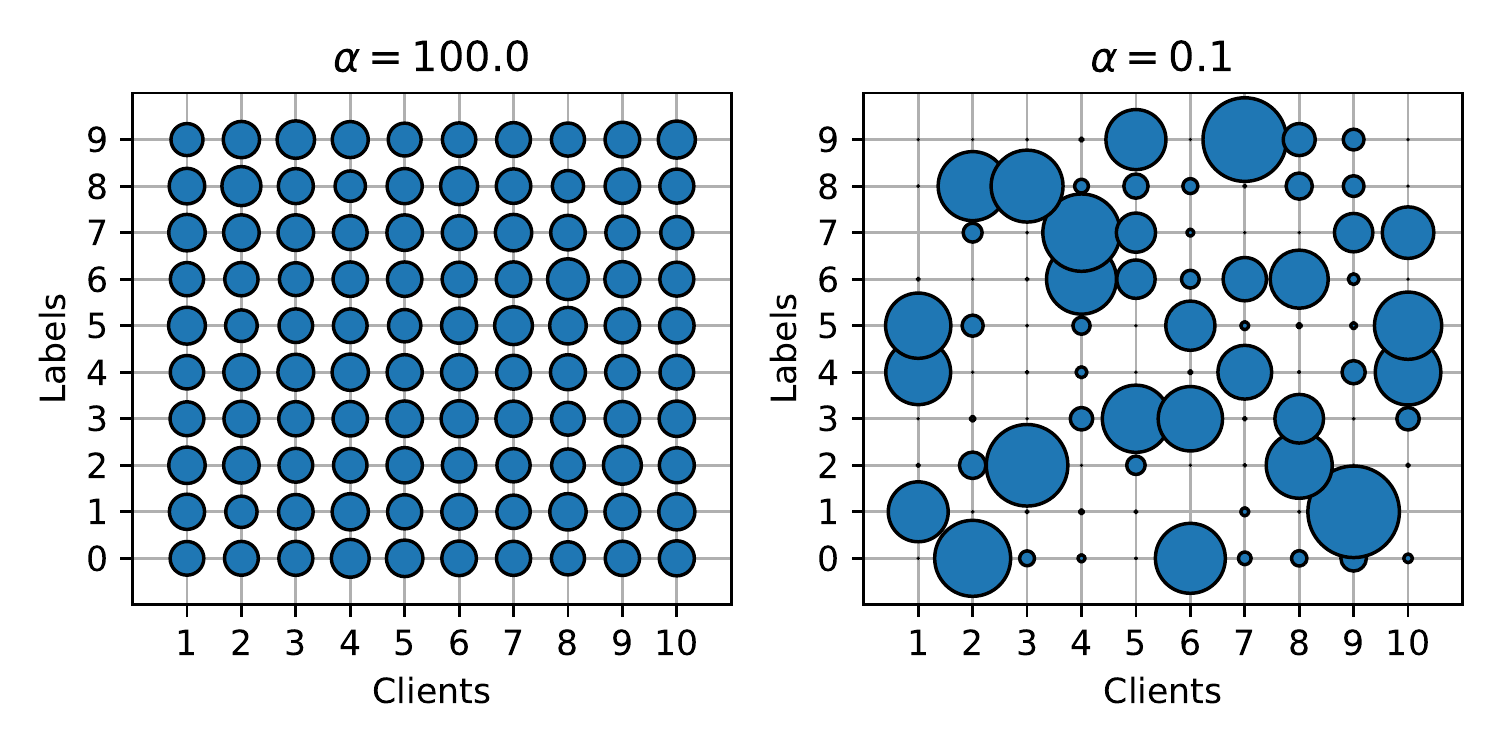}
    \caption{Illustration of the Dirichlet data splitting strategy we use throughout the paper, exemplary for a Federated Learning setting with 10 Clients and 10 different classes. Marker size indicates the number of samples held by one client for each particular class. Lower values of $\alpha$ lead to more heterogeneous distributions of client data. Figure adapted from \cite{lin2020ensemble}. }
    \label{fig:alpha1}
\end{figure}

\section{Investigating the communication Properties of Federated Distillation}
\label{sec3}
In this section we investigate the communication properties of Federated Distillation. The total amount of communication necessary to transfer the soft-label information in each round is given by the product of the distillation data set size and the average amount of bits required to store the value of one soft-label
\begin{align}
\label{eq:communication}
    \mathtt{b}_{total} = |X^{pub}| \times (H(Y_i) + \eta).
\end{align}
Hereby $H(Y_i)$ is the entropy of the soft-labels and $\eta$ indicates the coding inefficiency. In conventional Federated Distillation as proposed in \cite{jeong2018distill,itahara2020distill}, the soft-label information is stored at 32 bit floating point precision and thus we have
\begin{align}
    \mathtt{b}_{total} = |X^{pub}| \times \text{dim}(\mathcal{Y}) \times 32\text{bit}.
\end{align}
Following eq. \eqref{eq:communication}, a reduction of the communication overhead can be achieved by either
\begin{itemize}
    \item[(a)] reducing the size of the distillation dataset,
    \item[(b)] reducing the entropy of the soft-labels, or
    \item[(c)] improving the efficiency of the coding technique. 
\end{itemize}

In this section, we will look at all three of these determining factors and investigate their relative impact on the Federated Learning performance.

In the preliminary experiments performed in this section we consider Federated Learning settings with 20 clients among which we split the training data according to a Dirichlet distribution, as illustrated in Figure~\ref{fig:alpha1}. More details on the experiment setup can be found in section \ref{sec:experiments}.

\begin{figure}
    \centering
    \includegraphics[width=0.5\textwidth]{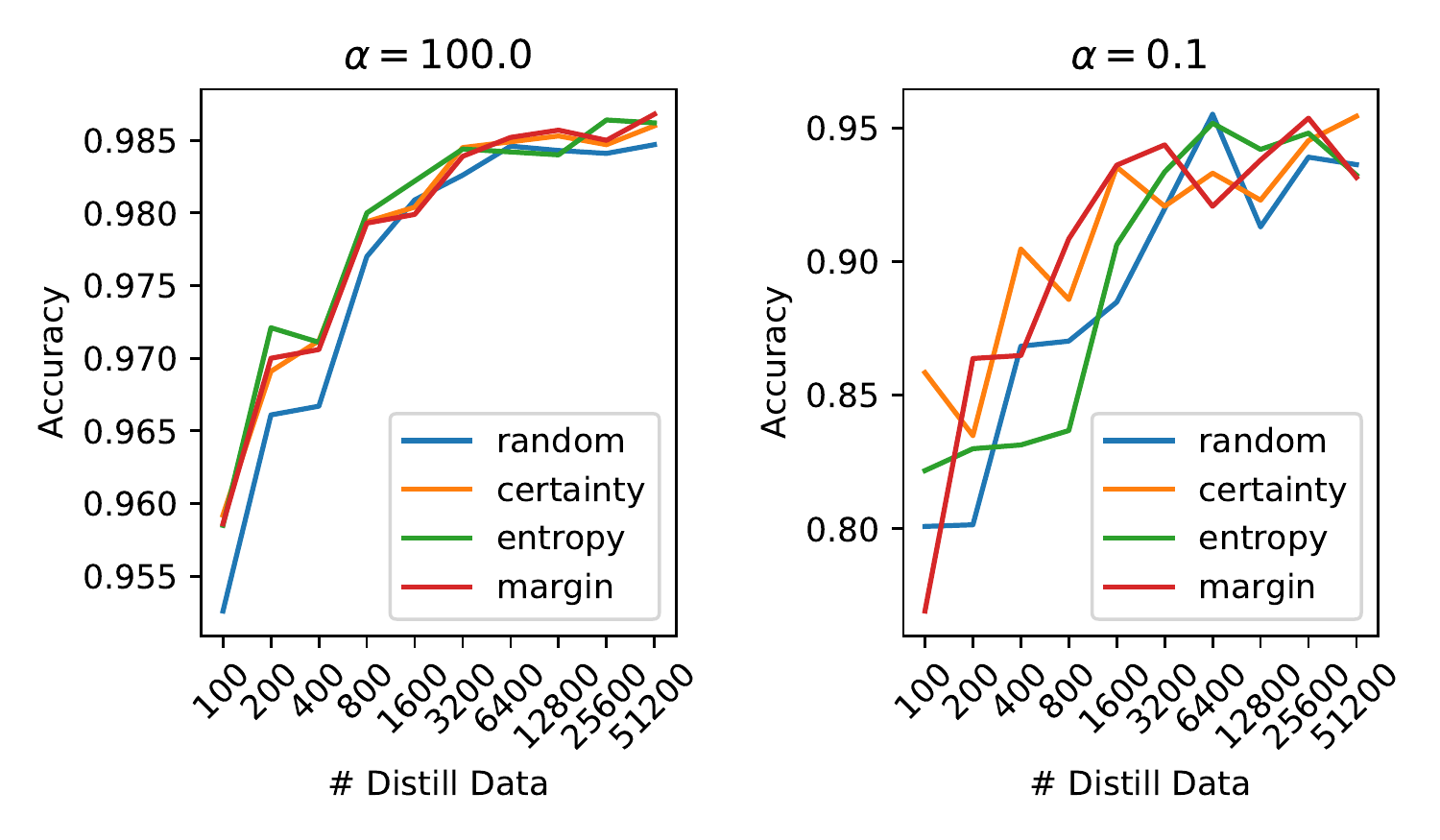}
    \caption{Effect of distillation dataset size with different (active) selection strategies using LeNet on MNIST.}
    \label{fig:active_mnist}
\end{figure}

\subsection{Size of Distillation Dataset}
\label{sec:datasize}
As the communication overhead in Federated Distillation is directly proportional to the number of data points used for distillation, restricting the size of the distillation data is the most straight-forward way to reduce communication. It is commonly known however, that in machine learning (and deep learning in particular), the size of the training data set has strong impact on the generalization capacity of any trained classifier \cite{vapnik2013nature}. The machine learning discipline of active learning has developed techniques to systematically select samples from a larger pool of data for training with the goal to achieve higher performance with fewer samples of data. Here, we adapt four popular active learning techniques to the setting of Federated Distillation and compare their performance when used to select distillation data sets of different sizes.  Let 
\begin{align}
    \text{top}_n[x \mapsto \Psi(x)] : \mathcal{D} \rightarrow \mathcal{D}
\end{align}
be the operator that maps a data set to one of its subsets of size $n$, by selecting the top $n$ elements according to the criterion $x \mapsto \Psi(x)$. Then we can define the "entropy", "certainty" and "margin" selection strategy as follows:

\begin{align}
    D^{entropy}_n &= \text{top}_n[x \mapsto H(f_{\theta}(x))](X^{pub})
    \end{align}
    \begin{align}
        D^{certainty}_n &= \text{top}_n[x \mapsto -\max(f_{\theta}(x))](X^{pub})
        \end{align}
        \begin{align}
            D^{margin}_n &= \text{top}_n[x \mapsto {\max}_2(f_{\theta}(x)) - \max(f_{\theta}(x))](X^{pub}) 
\end{align}
Hereby, $H(p)=-\sum_i p_i\log(p_i)$ denotes the entropy, $\max(p)$ represents the maximum value in the vector of probabilities $p$, and $\max_2(p)=\max(p\setminus \{\argmax(p)\})$ denotes the second-largest element of $p$. We also consider the selection strategy of picking $n$ data-points at random in each round.

In each communication round of Federated Distillation we select a subset of $n$ data points for distillation according to one of the above defined strategies based on the model $\theta$ which was used in the previous round. The results of this experiment are shown in Figure~\ref{fig:active_mnist}. As we can see, the performance of Federated Distillation strongly depends on the size of the distillation data set. The effect of using active learning strategies to systematically select data points on the other hand is rather low. While in the IID regime ($\alpha=100.0$) the active learning strategies slightly improve the Federated Distillation performance, the situation is rather unclear in the non-IID regime ($\alpha=0.1$). From this we conclude, that in most situations the performance gains obtained by using active learning strategies do not justify the additional computational overhead incurred by these techniques (evaluating $f_\theta(x)$ on the entire accessible distillation data). In the remainder of this manuscript we will thus restrict our analysis to randomly selected distillation data sets of fixed size.

\subsection{Soft-Label Quantization}
\label{sec:quantization}
\begin{figure}
    \centering
    \includegraphics[width=0.5\textwidth]{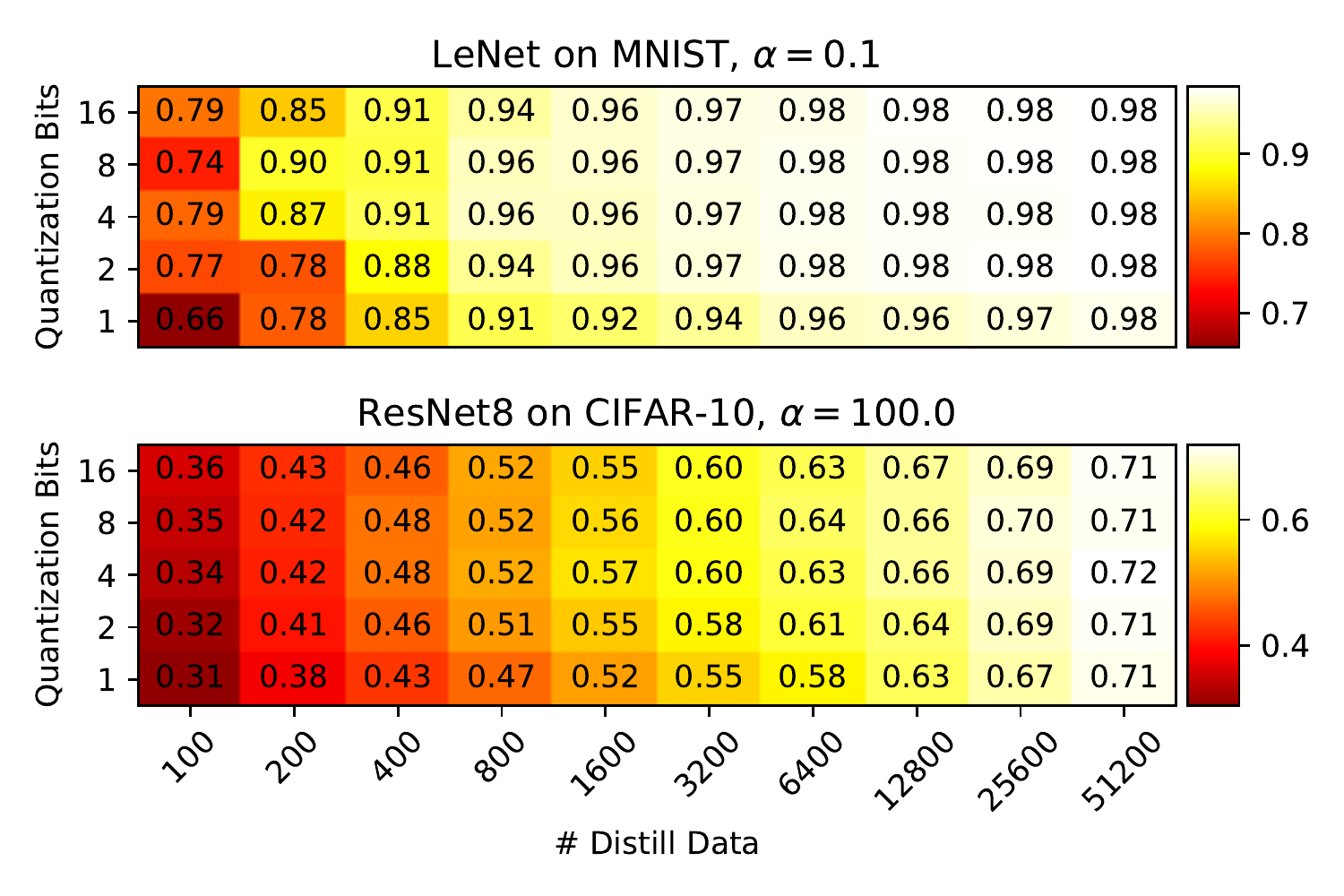}
    \caption{Effect of distillation data set size and quantization strength on training performance in Federated Distillation using LeNet on MNIST at $\alpha=0.1$ and ResNet-8 on CIFAR-10 at $\alpha=100.0$.}
    \label{fig:n_data_quant}
\end{figure}

Quantization is a popular technique to reduce communication and has been successfully applied in Federated Averaging to reduce the size of the parameter updates \cite{konevcny2016federated, sattler2018sparse, xu2020compress}. Quantization techniques, however, so far have not been applied to Federated Distillation. Here we consider constrained uniform quantization to reduce the entropy of the communicated soft-labels. Let $p\in \mathcal{Y}$ be a vector of soft-label probabilities. Then we obtain the quantized soft-label $q$ via constrained uniform quantization as follows

\begin{align}
\label{eq:quant}
    q = \mathcal{Q}_b(p) = \arg\min_{\stackrel{q_i\in\{\frac{l}{2^b-1}, l\in 0,..,2^b-1\}}{\sum_{i}q_i=1}} \|q-p\|_1
\end{align}

The optimization problem above can be solved in log-linear time. In case the optimization problem in \eqref{eq:quant} does not have a unique solution, we randomly break the tie.
As can be easily seen, for $b=1$, the quantization operator $\mathcal{Q}_b$ is equivalent to the maximum vote:
\begin{align}
\label{eq:onehot}
    \mathcal{Q}_1(p)_i = \begin{cases}
    1 & \text{if }i=\arg\max(p)\\
    0 & \text{ else}
    \end{cases}
\end{align}

Constrained uniform quantization as defined above reduces the number of bits required to communicate any vector of probabilities from $32\text{dim}(\mathcal{Y})$ to $b\text{dim}(\mathcal{Y})$.

Figure~\ref{fig:n_data_quant} shows the effect of different distillation data set sizes and quantization levels on the model accuracy after a fixed number of communication rounds. We observe two interesting trends. Firstly we observe that, while reducing the number of quantization bits by half has the same effect on the communication overhead as reducing the size of the distillation data by half, it has a much lower impact on training performance. This holds across all levels of quantization and all distillation data set sizes. Secondly, if distillation data is abundant (here $n\geq12800$), we observe that the harmful effect of quantization even reverses and for 51200 distillation datapoints, the highest performance is actually achieved at the highest compression level. While counter-intuitive at first glance, this effect could be attributed to the regularizing effect that quantization has on the training process.

These results indicate that as a means for reducing communication, quantization should be strictly preferred over distillation data set reduction, especially if one has access to a large distillation data set. In the following, we will thus concentrate our analysis on the strongest compression operator $\mathcal{Q}_1$.


\subsection{Efficient Encoding}
\label{sec:encoding}

\begin{figure}
    \centering
    \includegraphics[width=0.5\textwidth]{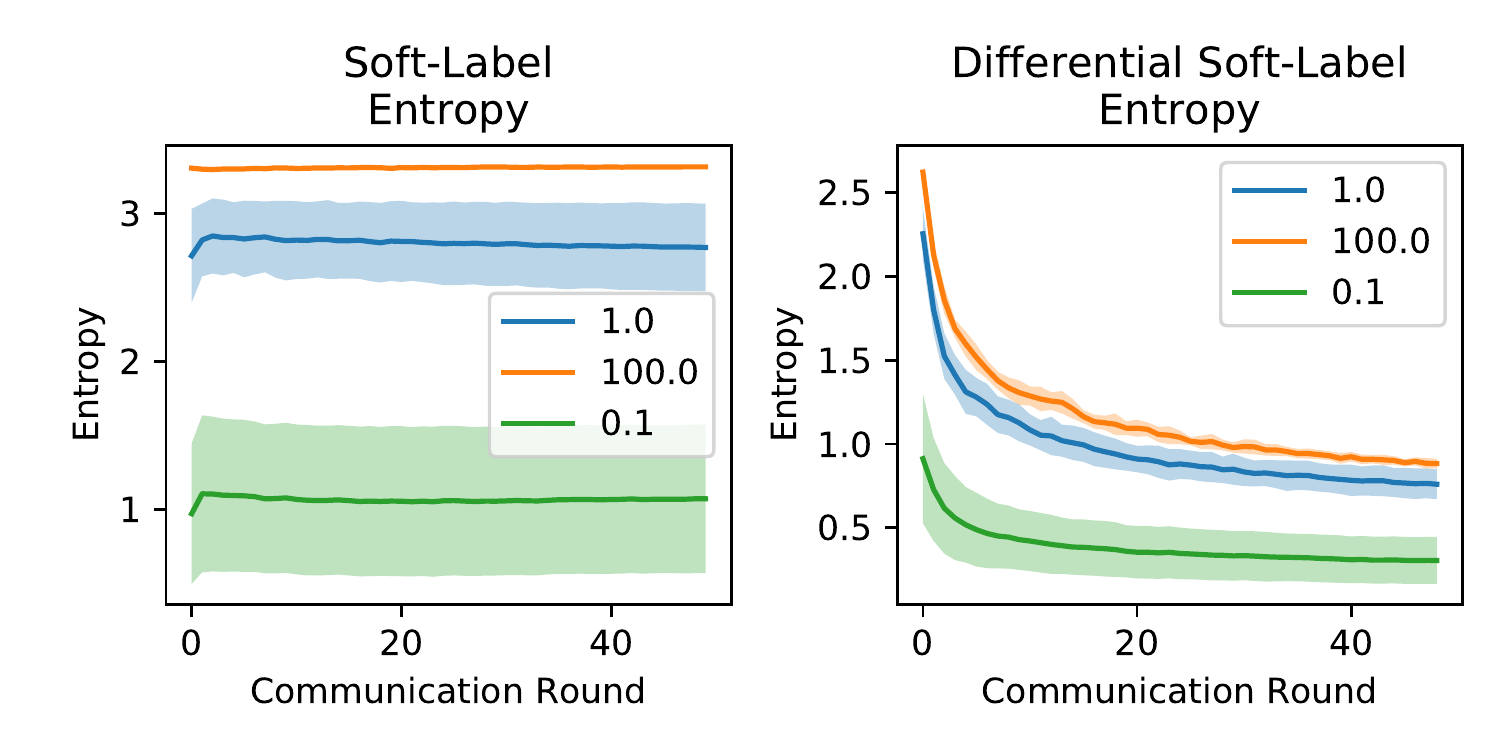}
    \caption{Communication over the course of training. When communicating compressed soft-labels directly communication stays constant over the course of training. When instead communicating differential soft-labels, communication steadily decreases. Results for ResNet-8 on CIFAR-10.}
    \label{fig:ent_diff_ent}
\end{figure}

In this section we investigate efficient lossless coding techniques to minimize the size of the compressed soft-label representations. As shown in eq.\ \eqref{eq:onehot}, applying the compression operator $\mathcal{Q}_1$ to a vector of probabilities $p$ results in a one-hot vector of size $\text{dim}(\mathcal{Y})$. As this one-hot vector can also be represented by an integer number between 1 and $\text{dim}(\mathcal{Y})$, a straight-forward encoding process would comprise of communicating 
\begin{align}
    \tilde{Y}_i = \{\mathcal{Q}_1(f_{\theta_i}(x))|x \in X^{pub}\}
\end{align}
as an array of $|X^{pub}|$ integer values, using up $|X^{pub}|\times \log_2(\text{dim}(\mathcal{Y}))$ bits of data in total.

\begin{figure}
    \centering
    \includegraphics[width=0.5\textwidth]{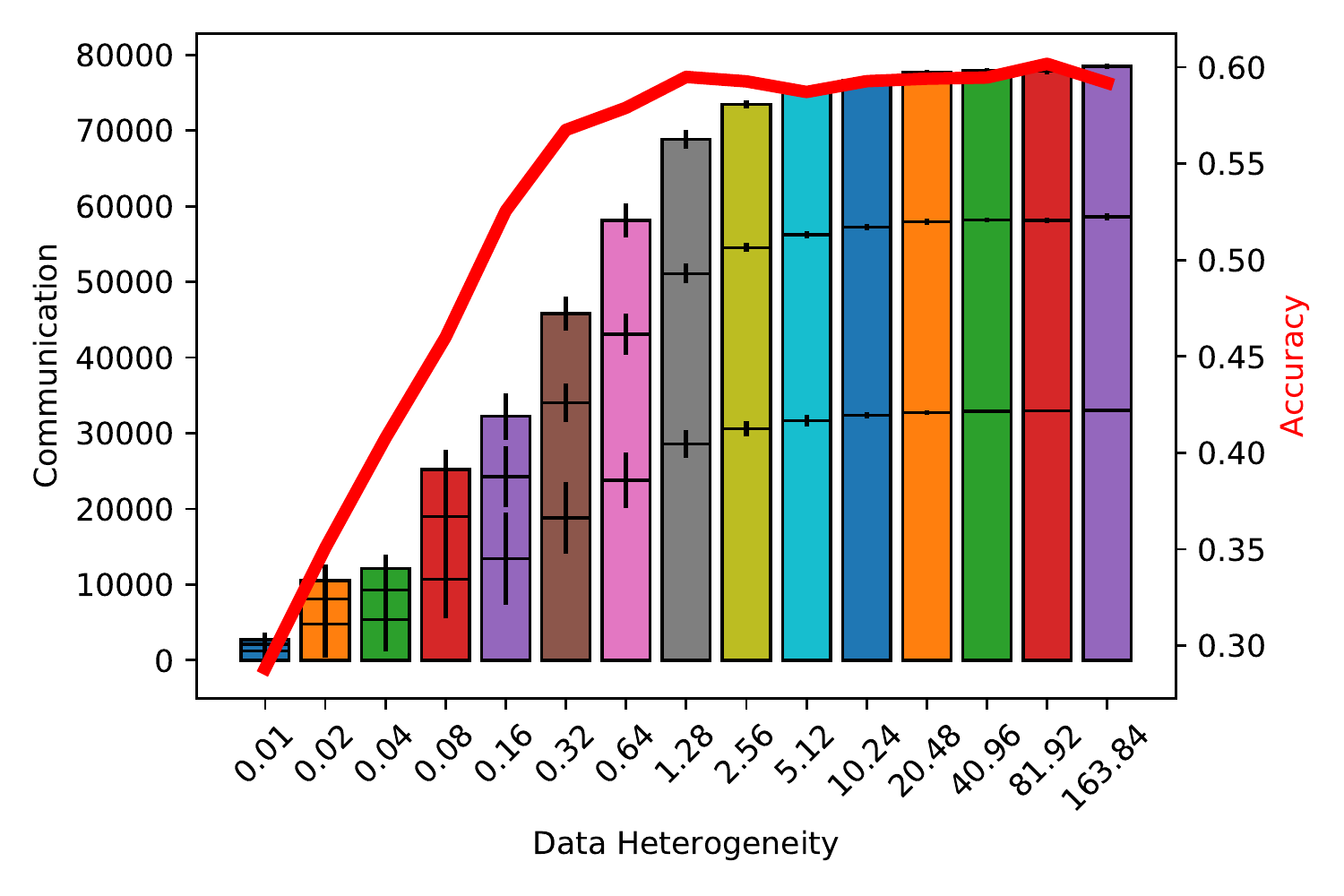}
    \caption{Upstream Communication and model accuracy at different levels of heterogeneity $\alpha$ for Resnet-8 trained on CIFAR. Communication varies by more than an order of magnitude between the most homogeneous and the most heterogeneous setting. }
    \label{fig:non_iidcommunication}
\end{figure}

This however is only an upper bound on the true entropy $H(\tilde{Y}_i)$, which highly depends on the distribution of max predictions in $\tilde{Y}_i$. Figure~\ref{fig:ent_diff_ent} (left) shows the development of $H(\tilde{Y}_i)$ over the course of 50 communication rounds for a Federated Learning problem with 20 clients training ResNet-8 on CIFAR-10 at different levels of data heterogeneity. As we can see, the true entropy is well below the theoretical maximum of $\log_2(10)$ (for CIFAR-10 we have $\text{dim}(\mathcal{Y})=10$) and decreases with increasing heterogeneity $\alpha$ down to around $H(\tilde{Y}_i)\approx 1$ at $\alpha=0.1$. This behaviour is expected, as the labels in the client training data, and consequently also their predictions $\tilde{Y}_i$, get more concentrated with increasing heterogeneity in the data.  

Additional knowledge about the distribution of $\tilde{Y}_i$ can be used to further reduce the entropy. Since Federated Distillation is empirically known to converge \cite{}\cite{}, we formulate the hypothesis that, given a fixed distillation data set, there should be a relatively large overlap between the predictions made by a client in the current round $T$ and those made in the previous round $T-1$. Furthermore, as the Federated Distillation process converges toward a stationary solution, this overlap is expected to grow bigger over time. 

High agreement between consecutive data points in a stream of data is a phenomenon commonly encountered in communication. The effect for instance can also be found in video data, where consecutive frames are often highly correlated. The canonical technique to exploit this pattern is differential coding (resp. delta coding or predictive coding) \cite{sayood2017datacompression}, which relies on only communicating "new" information in order to achieve higher compression rates.

To test our hypothesis, we apply delta coding to the quantized predictions of two consecutive rounds $\tilde{Y}^t$ and $\tilde{Y}^{t-1}$ by setting 
\begin{align}
\label{eq:delta}
(\hat{Y}^t)_l = \begin{cases}
(\tilde{Y}^t)_l & \text{if }(\tilde{Y}^t)_l\neq (\tilde{Y}^{t-1})_l \\
0 & \text{else}
\end{cases} ~~\forall l
\end{align}
and measuring the entropy (in slight abuse of notation this assumes an arbitrary but fixed ordering of the set $\tilde{Y}$ and the same distillation data set $X^{pub}$ to be used in all rounds). It should be noted, that all of the information contained in $\tilde{Y}^t$ can be retained from $\hat{Y}^t$ by comparing with the previous message $\hat{Y}^{t-1}$. This only requires minor additional bookkeeping by the central server (which is typically assumed to have access to strong computational resources).

Figure~\ref{fig:ent_diff_ent} (right) shows the development of the entropy of the differential updates $H(\hat{Y}_i)$. As we can see, the differential soft-label entropy behaves exactly as predicted and $H(\hat{Y}_i)$ is lower than $H(\tilde{Y}_i)$ from the first round on and smoothly decreases over the course of training. We note that, curiously, the development of the differential soft-label entropy over time can be very accurately predicted via the functional relation $H(\hat{Y}^t)\approx c t^{-d}$ for some constants $c,d$. We were able to replicate this behaviour across different model architectures and Federated Learning settings, hinting at an interesting underlying mathematical relationship, which could be the subject of future studies. 

Figure~\ref{fig:non_iidcommunication} explores in more detail the influence of data heterogeneity on the amount of communication. It displays the upstream communication in the first three rounds of Federated Distillation with quantization and differential soft-label encoding. The resulting model accuracy is also given (indicated by the red curve). As we can see, the amount of communication monotonically decreases when lowering the value of $\alpha$ (thus increasing the heterogeneity), with more than an order of magnitude difference between the most homogeneous and the most heterogeneous setting. This suggest, that using the proposed quantization and encoding scheme is particularly beneficial in heterogeneous settings.

\begin{figure}
    \centering
    \includegraphics[width=0.5\textwidth]{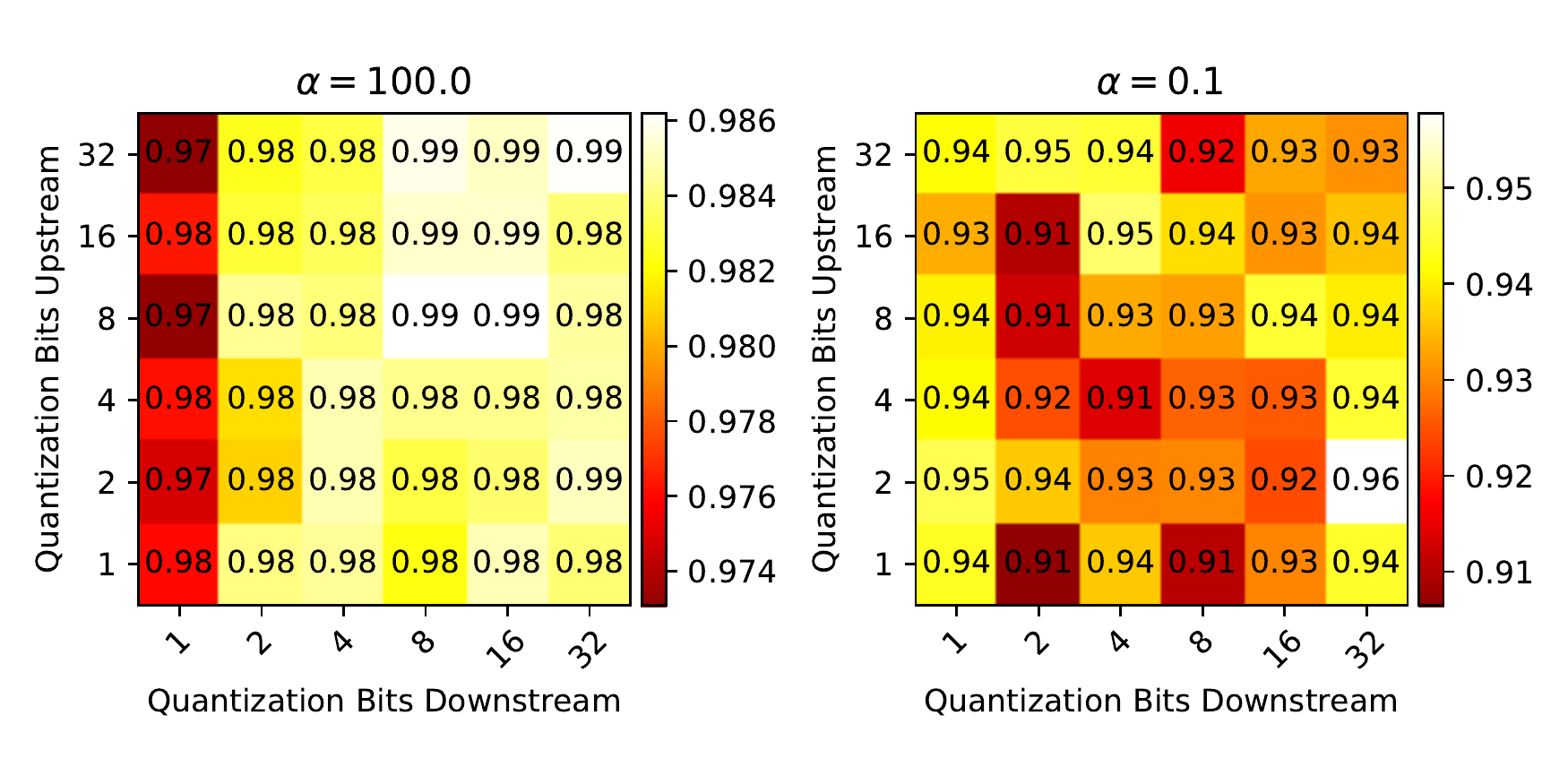}
    \caption{Effect of different levels of upstream and downstream quantization on the training performance. LeNet on MNIST at two different level of heterogeneity $\alpha$. Displayed is the maximum accuracy achieved after 20 communication rounds.}
    \label{fig:mnist_downstream}
\end{figure}

\subsection{Efficient Downstream Communication}
\label{sec:downstream}

\begin{table}[]
    \centering
        \caption{Effect of the initialization in Federated Distillation on the maximum Accuracy achieved after 20 communication rounds. Display are mean and standard deviation over 10 runs of Federated Distillation with a client participation rate of 0.4. Our proposed dual distillation approach closes the gap between random initialization and initialization from previous. }
        \label{tab:init}
\begin{tabular}{llllll}
\toprule
 &   &       &             Init Random &             Init Prev. &             Dual Distill \\
&  & $\alpha$ &               &               &               \\
\midrule

LeNet &  & 100.0 &  0.977(0.001) &  \textbf{0.985(0.001)} &  0.981(0.001) \\
&     & 1.0   &  0.974(0.001) &  \textbf{0.982(0.001)} &  0.980(0.000) \\
&     & 0.1   &  0.920(0.005) &  \textbf{0.944(0.008)} &  0.915(0.005) \\
\midrule
ResNet & & 100.0 &  0.763(0.002) &  0.704(0.003) &  \textbf{0.776(0.003)} \\
&     & 1.0   &  0.738(0.002) &  0.691(0.001) &  \textbf{0.752(0.003)} \\
&     & 0.1   &  0.473(0.027) &  \textbf{0.481(0.007)} &  0.474(0.021) \\
\bottomrule
\end{tabular}
\end{table}

So far we have only considered the upstream communication from the clients to the server. While in most Federated Learning settings with mobile and IoT devices the up-link channel is more strongly constrained than the down-link, it is still desirable to also reduce the downstream communication as much as possible. 

One issue that typically arises in compressed Federated Learning, is that clients may run out of sync if the participation rate is below 100\%. This is because any state information (like the model state $\theta$) becomes stale if clients do not participate in every round. To keep the client models synchronized if the participation rate is below 100\% they need to either download the latest master-model $\theta$ from the server in every round (resulting in high downstream communication) or alternatively they can also be randomly re-initialized in every round. 

To illustrate this point, Table \ref{tab:init} shows the maximum accuracy achieved after 20 communication rounds of Federated Distillation with three different client initialization schemes and three different neural networks at varying levels of data heterogeneity. As we can see, initializing the client models randomly before distillation achieves worse performance than using the distilled model from the previous round as initialization point. 

To close the performance gap, we propose a novel dual distillation technique, which avoids de-synchronization of client models at arbitrary participation rates. In dual distillation, instead of directly sending the aggregated soft-labels
\begin{align}
    Y^{pub} = \frac{1}{|I_t|}\sum_{i\in I_t}\tilde{Y}^{pub}_i
\end{align}
to the clients, the server of it's own first performs a distillation step
\begin{align}
    \theta^t_S \leftarrow \text{train}(\theta^{t-1}_S, X^{pub}, Y^{pub})
\end{align}
using the model $\theta^{t-1}_S$ which was distilled in the previous round as initialization point. This way the training information stored in $\theta^{t-1}_S$ is not lost.
Then, the server computes soft-labels using the newly distilled model,
\begin{align}
    Y^{pub}_S = \{f_{\theta_S}(x)|x \in X^{pub}\}
\end{align}
 and sends them to the clients.
Starting from a random initialization, the participating clients then distill from the server predictions to mimic the server model
\begin{align}
    \theta \leftarrow \text{train}(\theta_0, X^{pub}, Y^{pub}_S)
\end{align}
This way the clients are indirectly initialized with all the accumulated training information stored in $\theta_S$, before going into the next round of local training.

This allows us now to communicate soft-labels in upstream and downstream and appreciate the resulting communication savings in both directions. To further reduce the amount of downstream communication, we can also quantize the server soft-labels $Y^{pub}_S$ before communication, using the same constrained compression operator $\mathcal{Q}_{b_{down}}$ that we used in the upstream. 

Figure \ref{fig:mnist_downstream} shows the effects of different levels of upstream and downstream quantization on the training performance of LeNet trained on MNIST using Federated Distillation after 20 communication rounds. As we can see in the IID setting with $\alpha=100.0$, downstream quantization appears to have a slightly stronger effect on the model performance, with a maximum accuracy drop of 1\% at the strongest quantization level at $b_{down}=1$. In the more heterogeneous setting with $\alpha=0.1$ it is more difficult to observe such a trend. Here, the strongest level of upstream and downstream compression outperforms the uncompressed FD. It appears that using quantization in both upstream and downstream is a very promising technique for reducing communication.

\begin{algorithm}[t]
\caption{Compressed Federated Distillation}\label{alg:CFD}
\label{alg:1}
\DontPrintSemicolon
\textbf{init:} Set upstream and downstream precision $b_{up}$ and $b_{down}$. Every Client holds a different local data set $D_i=(X_i, Y_i)$ as well as the common public data set $X^{pub}$, with size $|X^{pub}|=n$. \\
\For{$t=1,..,T$}{
\For{$i \in I_t\subseteq \{1,..,\textnormal{[Number of Clients]}\}$ \textbf{in parallel}}{
\underline{Client $C_i$ does:}\\
\textbullet~$\theta\leftarrow\text{random\_init}()$ \hfill \# Initialize\\
\If{$t>1$}{
\textbullet~$\text{download}_{S\rightarrow C_i}(\tilde{Y}^{pub}_{S})$\\
\textbullet~$\theta\leftarrow \text{train}(\theta, X^{pub}, \tilde{Y}^{pub}_{S})$\hfill \# Distillation\\
}
\textbullet~$\theta_i\leftarrow \text{train}(\theta, X_i, Y_i)$\hfill \# Local Training\\
\textbullet~$Y^{pub}_i\leftarrow f_{\theta_i}(X^{pub})$\hfill\# Compute Soft-Labels\\
\textbullet~$\tilde{Y}^{pub}_i\leftarrow \mathcal{Q}_{b_{up}}(Y^{pub}_i)$\hfill\# Compress Soft-Labels\\
\textbullet~$\text{upload}_{C_i \rightarrow S}(\tilde{Y}^{pub}_i)$ \hfill \# Upload
}
\underline{Server $S$ does:}\\
\textbullet~$Y^{pub}\leftarrow \frac{1}{|I_t|}\sum_{i\in I_t}\tilde{Y}^{pub}_i$ \hfill \# Aggregate\\
\textbullet~$\theta_{S}\leftarrow \text{train}(\theta_S, X^{pub}, Y^{pub})$\hfill \# Server Distillation\\
\textbullet~$Y^{pub}_{S}\leftarrow f_{\theta_{S}}(X^{pub})$\hfill\# Compute Soft-Labels\\
\textbullet~$\tilde{Y}^{pub}_{S}\leftarrow \mathcal{Q}_{b_{down}}(Y^{pub}_{S})$\hfill\# Compress Soft-Labels\\
}
\textbf{return} $\theta_S$
\end{algorithm}

\section{Compressed Federated Distillation}
\label{sec4}
\label{section:compressed_fed_distillation}

\begin{figure}
    \centering
    \includegraphics[width=0.5\textwidth]{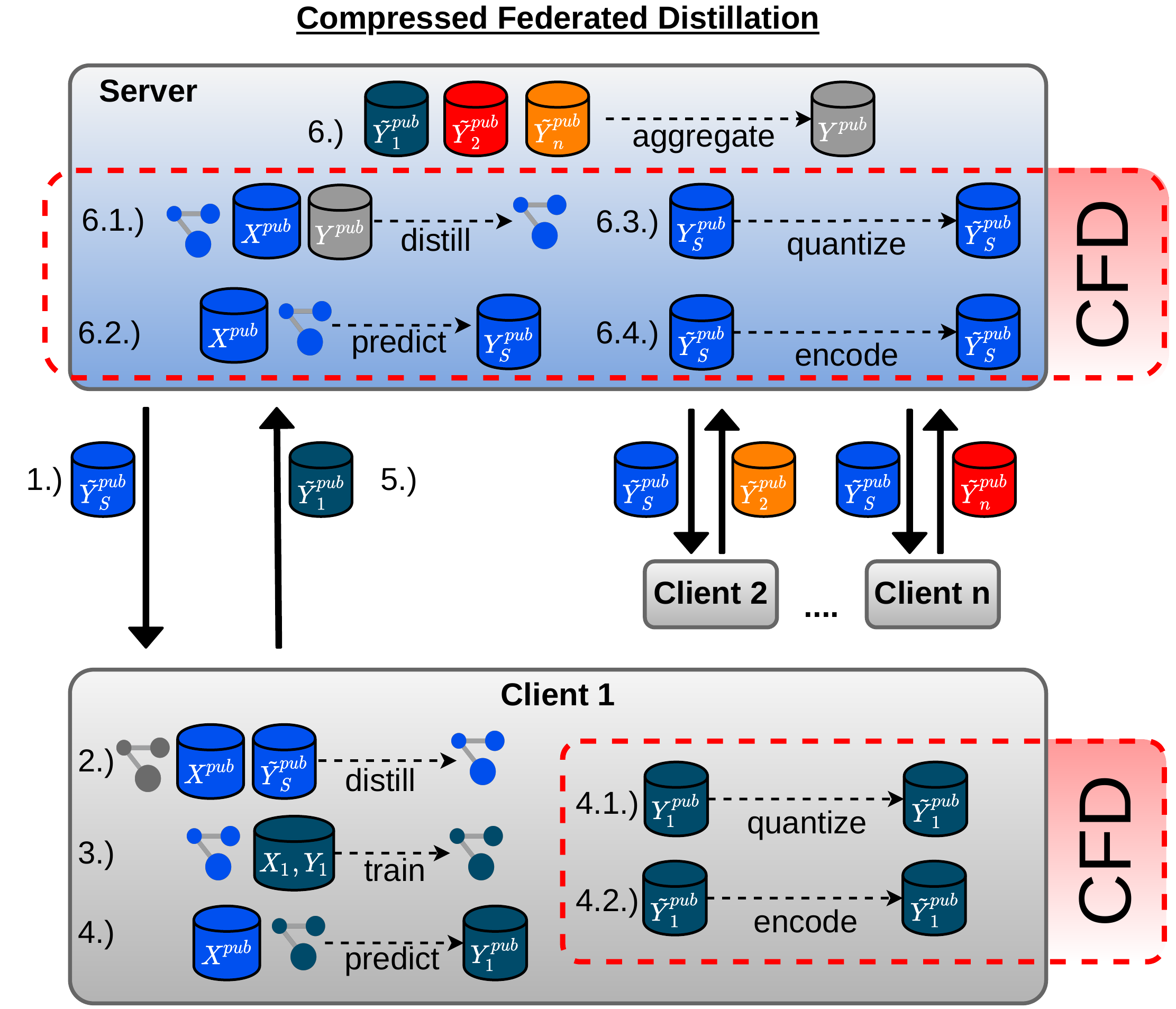}
    \caption{Our proposed Compressed Federated Distillation method employs distill data curation (see Sec. \ref{sec:datasize}), soft-label quantization (4.2., see Sec \ref{sec:quantization}) and delta-coding (4.3., see Sec. \ref{sec:encoding}) to minimize the communication from the clients to the server. Furthermore, CFD uses dual distillation (6.1., 6.2., see Sec. \ref{sec:downstream}) to keep clients synchronized in situations when full client participation in every round can not be ensured. On top of that CFD also uses quantization (6.3.) and delta-coding (6.4.) in the downstream, to reduce the communication from the server to the clients.}
    \label{fig:contrib}
\end{figure}

In this section, we combine the insights of the previous section and propose Compressed Federated Distillation (CFD). CFD extends the conventional Federated Distillation framework by the following five techniques:
\begin{enumerate}
    \item \textbf{Distill Data curation (Alg \ref{alg:CFD} - {\scriptsize 1})}: We select a fixed random subset $X^{pub}$ of the available distillation data for training, this subset is not varied over the course of training.
    \item \textbf{Upstream quantization (Alg \ref{alg:CFD} - {\scriptsize 12})}:  We reduce the bit-width of the client soft-labels by applying the constrained quantization operator $\mathcal{Q}$ (eq. \eqref{eq:quant}).
    \item \textbf{Delta coding (Alg \ref{alg:CFD} - {\scriptsize 12})}: The quantized soft-label are encoded using an efficient arithmetic entropy coding technique, like CABAC \cite{marpe2003context}.  Additionally, we use delta coding (eq. \eqref{eq:delta}), to further reduce the entropy of the quantized soft-label information $\tilde{Y}_i$.
    \item \textbf{Dual Distillation (Alg \ref{alg:CFD} - {\scriptsize 17, 18})}: In every round, we distill a server model $\theta_S$ from the aggregated soft-labels. This server model accumulates training information from all previous communication rounds. The clients are then trained to match the predictions of this server model. This avoids loss of information in settings where clients do not participate in every round.
    \item \textbf{Downstream Quantization (Alg \ref{alg:CFD} - {\scriptsize 19})} We apply constrained quantization $\mathcal{Q}$ also to the predictions of the server model before sending them down to the clients. The clients then, starting from a random initialization, are trained to mimic the predictions of the server model.
\end{enumerate}
The training procedure is illustrated in Figure~\ref{fig:contrib} and formally described in Algorithm \ref{alg:CFD}.

The performance of our algorithm in every round $t$ is determined from the distilled model $\theta_S$ on the validation data set.

\section{Experiments}
\label{sec5}
\label{sec:experiments}
In this section we empirically evaluate our proposed Compressed Federated Distillation method and compare its performance against the natural baselines of Federated Averaging \cite{mcmahan2017communication} and Federated Distillation \cite{itahara2020distill}. The experimental setup is given as follows:


\begin{table*}[]
    \caption{Upstream and downstream communication in [MB] necessary to achieve accuracy targets in Federated Learning on the CIFAR-10 data set with different neural network models and at different levels of data heterogeneity. In all scenarios, our proposed CFD method outperforms the baselines by a wide margin. Federated Learning setting with 20 clients and 0.4 participation rate. For distillation methods 80000 data point from the STL-10 data set were used. A value of "n.a." signifies that the method did not achieve the target accuracy within 50 communication rounds.}
    \label{tab:results_image}
    \centering
\begin{tabular}{llllllllll}
\toprule
      &      &     &      &        FA &      FD & CFD-1-32 & CFD$_\Delta$-1-32 & CFD-1-1 & CFD$_\Delta$-1-1 \\
Model & Target-Acc & $\alpha$ & Up/Down &           &         &          &               &         &              \\
\midrule
Alexnet & 0.68 & 100.0 & up &     n.a. &  89.60 &     0.94 &          \textbf{0.74} &    n.a. &         n.a. \\
      &      &     & down &     n.a. &  \textbf{89.60} &    92.80 &         92.80 &    n.a. &         n.a. \\
      & 0.64 & 1.0 & up &     n.a. &  38.40 &     0.61 &          \textbf{0.49} &    0.76 &         0.62 \\
      &      &     & down &     n.a. &  38.40 &    67.20 &         67.20 &    0.84 &         \textbf{0.42} \\
      & 0.44 & 0.1 & up &     n.a. &   6.40 &     0.09 &          \textbf{0.08} &    0.11 &         0.10 \\
      &      &     & down &     n.a. &   6.40 &    19.20 &         19.20 &    0.17 &         \textbf{0.15} \\
      \midrule
Resnet18 & 0.71 & 100.0 & up &   760.35 &  44.80 &     0.56 &          \textbf{0.40} &    1.36 &         0.82 \\
      &      &     & down &   760.35 &  44.80 &    54.40 &         54.40 &    1.36 &         \textbf{0.39} \\
      & 0.68 & 1.0 & up &  1028.71 &  48.00 &     0.37 &          \textbf{0.28} &    0.64 &         0.43 \\
      &      &     & down &  1028.71 &  48.00 &    41.60 &         41.60 &    0.72 &         \textbf{0.34} \\
      & 0.45 & 0.1 & up &  1520.70 &  16.00 &     0.09 &          \textbf{0.08} &    0.52 &         0.40 \\
      &      &     & down &  1520.70 &  16.00 &    22.40 &         22.40 &    0.99 &         \textbf{0.92} \\
      \midrule
Vgg16 & 0.8 & 100.0 & up &   671.16 &  32.00 &     0.40 &          \textbf{0.29} &    0.76 &         0.47 \\
      &      &     & down &   671.16 &  32.00 &    38.40 &         38.40 &    0.76 &         \textbf{0.24} \\
      & 0.78 & 1.0 & up &  1281.30 &  28.80 &     0.38 &          \textbf{0.28} &    0.56 &         0.37 \\
      &      &     & down &  1281.30 &  28.80 &    41.60 &         41.60 &    0.62 &         \textbf{0.27} \\
      & 0.48 & 0.1 & up &  2928.69 &  25.60 &     0.11 &          \textbf{0.09} &    0.43 &         0.35 \\
      &      &     & down &  2928.69 &  25.60 &    28.80 &         28.80 &    0.77 &         \textbf{0.75} \\
\bottomrule
\end{tabular}
\end{table*}

\begin{figure*}
    \centering
    \includegraphics[width=\textwidth]{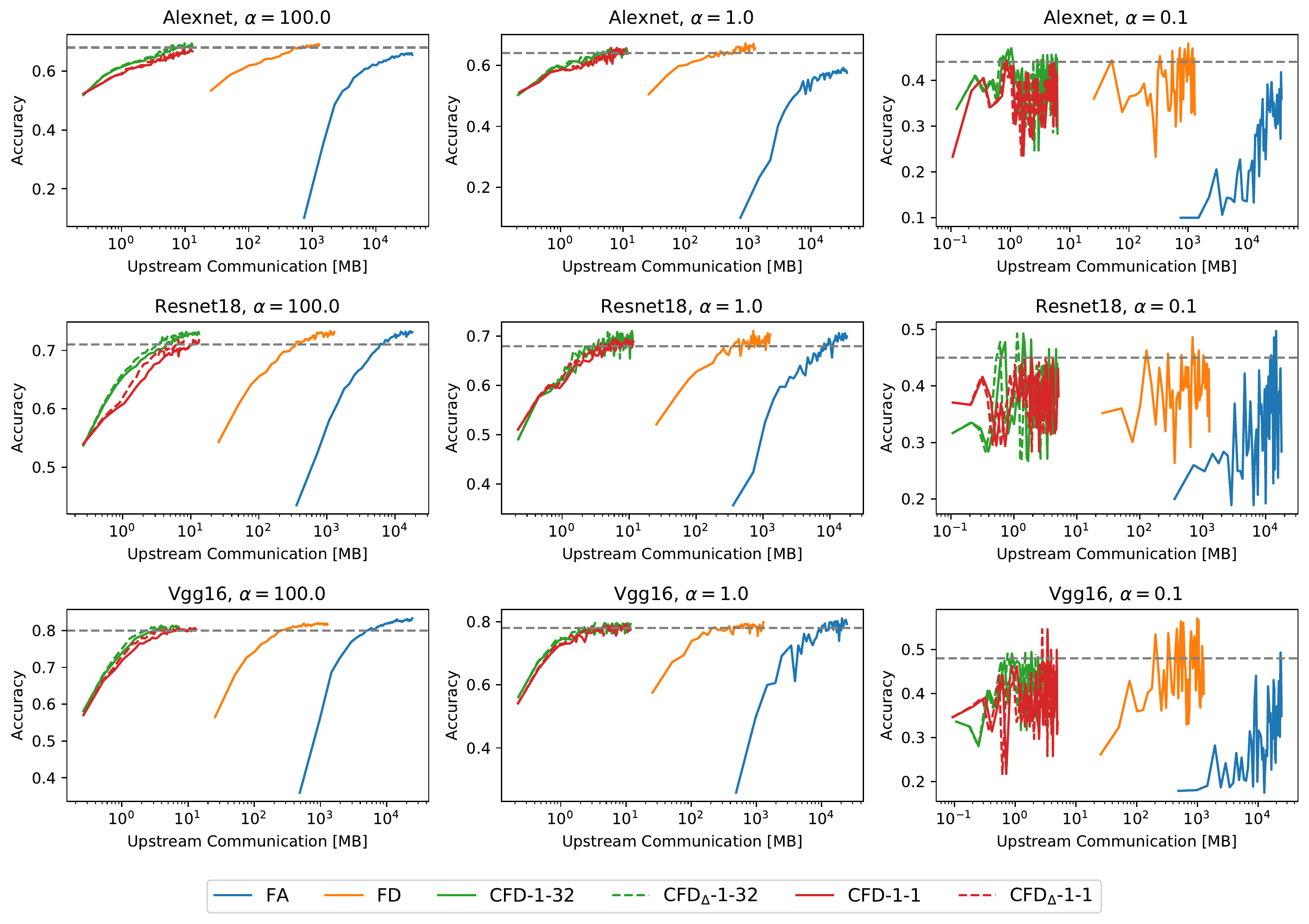}
    \caption{Model performance as a function of communicated bits for our proposed CFD method and baselines FA and FD, in Federated Learning on CIFAR-10 with different neural network models and at different levels of data heterogeneity. Federated Learning setting with 20 clients and 0.4 participation rate. For all distillation methods, 80000 data point from the STL-10 data set were used. }
    \label{fig:image_classifier_results_1}
\end{figure*}



\textbf{Data sets and models:} We evaluate CFD on both federated image and text classification problems with large scale convolutional and transformer neural networks, respectively. For our image classification problems we experiment with the following combinations of client- and/ distillation data: (MNIST / EMNIST \cite{cohen2017emnist}) and (CIFAR-10 / STL-10 \cite{coates2011analysis}). In both cases the distribution of the distillation data deviates from the one of the client data, as it would in realistic Federated Learning scenarios (MNIST contains handwritten digits, EMNIST contains handwritten characters, CIFAR-10 and STL-10 both contain different types of natural images). For our text classification problems we use disjoint splits of the SST2~\cite{socher2013recursive:sst2_dataset} and AG-News~\cite{zhang2015character:agnews_dataset} datasets for client training, distillation, and validation, respectively. We train LeNet- \cite{lecun1989handwritten}, VGG-type \cite{simonyan2014very}, AlexNet-type \cite{krizhevsky2012imagenet} and ResNet-type \cite{he2016deep} architectures with and without batch-normalization layers. The Alexnet, ResNet-18 and VGG-16 models used in our experiments contain 23.2M, 11.1M, and 15.2M parameters respectively. For our text classification experiments we fine-tune DistilBERT \cite{sanh2019distilbert}, a popular transformer model.  \\

\textbf{Federated learning environment and data partitioning}: For image classification problems, we consider Federated Learning settings with 20 clients. In all experiments, we split the training data evenly among the clients according to a dirichlet distribution following the procedure outlined in \cite{hsu2019measuring}. This allows us to smoothly adapt the level non-iid-ness in the client data using the dirichlet parameter $\alpha$. We experiment with values for $\alpha$ varying between 100.0 and 0.01. A value of $\alpha=100.0$ results in an almost identical label distributions, while setting $\alpha=0.01$ results in a split, where the vast majority of data on every client stems from one single class (see Figure~\ref{fig:alpha1} for an illustration). For image classifiers, we vary the client participation rate (in every round) between 40\% and 100\% and train for 50 communication rounds. For language models we set the number of clients to 10 and the participation rate to 100\% and train for 10 communication rounds. As it is standard convention in FL, the validation data follows the distribution of the clients' training data (not that of the distillation data). 

\textbf{Optimization details}: For the sake of simplicity, in all image classification tasks, we use the popular Adam \cite{kingma2014adam} optimizer with a fixed learning rate of $0.001$ across all baselines and for both the distillation and training on local private data. While a dedicated selection of optimizer and optimization hyperparameters might improve performance, our goal here is to give a fair comparison between the different Federated Learning algorithms. For language models, we perform one epoch of distillation with Adam, using a learning rate of $1\times10^{-5}$ and no weight decay. The clients' models are trained over one epoch with SGD in both scenarios, Federated Distillation and Federated Averaging, by setting the learning rate and momentum to 0.001 and 0.9, respectively. 

\textbf{Methods Compared}: We compare the performance of our method, Compressed Federated Distillation (CFD) including the variant with soft label encoding (referred in the experimental results as $\text{CFD}_\Delta$), with respect to the two natural baselines: Federated Averaging (FA) \cite{mcmahan2017communication} and Federated Distillation (FD) \cite{itahara2020distill}. For CFD, we test two configurations: For CFD-1-32 we only quantize the upstream communication by setting $b_{up}=1$ and $b_{down}=32$. For CFD-1-1 we quantize both the upstream and downstream communication and set $b_{up}=1$ and $b_{down}=1$. We also investigate the effects of using delta coding (as described in section \ref{sec:encoding}). CFD methods that use delta coding are indicated by CFD$_\Delta$.

\textbf{Measuring Communication}:
For the Baseline FD as well as our methods CFD and CFD$_\Delta$ we only measure the communication of soft-labels $Y^{pub}$. We explicitly ignore the communication cost of transferring the unlabeled public data set $X^{pub}$ to the participating clients. While clients technically need to download this data at one time before training, it is not part of the federated learning process. In communication sensitive applications, $X^{pub}$ could already be stored on the devices long before federated training is initiated and thus the timing of it's communication is much less critical. Other work \cite{lin2020ensemble} also demonstrates that $X^{pub}$ can be automatically generated on the clients using Generative Adversarial Networks.   

\subsection{Image classification results.}
We first investigate the communication properties of CFD on image classification benchmarks. Table \ref{tab:results_image} shows the amount of upstream and downstream bits required to achieve fixed accuracy targets for Alexnet, ResNet-18 and VGG-16 on CIFAR-10 at different levels of data heterogeneity between the clients. The corresponding training curves are given in Figure~\ref{fig:image_classifier_results_1}. As we can see, CFD is drastically more communication-efficient than the baselines FA and FD in all tested scenarios. For instance, for VGG-16 and $\alpha=100.0$, CFD$_\Delta$-1-1 achieves a target accuracy of 80\% by cumulatively communicating only 0.47~MB on average from the clients to the server and only 0.24~MB on average from the server to the clients. This is particularly remarkable, as one single transfer of the parameters of VGG-16 already takes up 61.01~MB. To achieve the same 80\% accuracy target, FA requires 671.16~MB of cumulative communication in both the upstream and the downstream, translating to more than three orders of magnitude in communication savings for CFD. When directly comparing with FD, which requires 32.00~MB, CFD still reduces the communication by about two orders of magnitude. Similar results can be observed for the two other tested neural networks ResNet-18 and Alexnet. On Alexnet FA even underperforms CFD w.r.t. to the maximum achieved accuracy and misses the accuracy target of 68\%. 

The communication savings are even larger in the non-iid settings with $\alpha=0.1$, where FA is known to perform poorly \cite{sattler2019robust}. For instance, when training ResNet-18 at $\alpha=0.1$ FA requires 1520.70 MB to achieve the accuracy target of 45\%. CFD$_\Delta$-1-32 requires only 0.08~MB to achieve the same accuracy, corresponding to a reduction in communication by a factor of $\times$ 19943. 

In all investigated settings, CFD$_\Delta$ methods that use delta coding are more efficient than those that do not. For instance for VGG-16 and $\alpha=1.0$ delta coding can bring down the cumulative upstream communication required to achieve 78\% accuracy from 0.38~MB to 0.28~MB for CFD-1-32. On the same benchmark delta coding also reduces the cumulative downstream communication from 0.56~MB to 0.37~MB for CFD-1-1.  

As can be seen in Figure~\ref{fig:image_classifier_results_1}, the heavily compressed CFD can keep up with the uncompressed baselines FD and FA w.r.t. maximum achieved accuracy on most benchmarks.


\input{lang_model_results/table_distilbert_bits_to_target}

\input{lang_model_results/fig_distilbert_results_sst2_and_agnews_100}

\subsection{Language model results.}Figure~\ref{fig:distilbert_results_sst2_and_agnews_100} shows the upstream communication for the different methods on a Federated fine-tuning task of DistilBERT, on the SST2 and AG-News data sets. In these experiments, we consider a Federated Learning setting with 10 clients, 100\% participation rate, and total of 10 communication rounds. We can highlight several important observations. First, while FA tends to achieve higher total accuracy than the other methods, it also requires several orders of magnitude more upstream communication. FD reduces the communication overhead with respect to FA by $\times$992 and $\times$279 in SST2 and AG-News datasets, respectively, at the expense of no more than 1\% accuracy degradation. CFD-1-32 (i.e., 1-bit upstream/32-bit downstream communication) and CFD-1-1 (i.e., 1-bit upstream/downstream communication) stand out as the most efficient techniques, resulting in communication savings when compared to FA of up to \emph{$\times$67000} in the SST2 dataset and $\times$17000 in the AG-News dataset, with negligible accuracy degradation. Fourth, we notice that in this particular set of experiments, soft label encoding (see $\text{CFD}_\Delta$-1-1 and $\text{CFD}_\Delta$-1-32), slightly increases the communication overhead with respect to CFD (in both cases, CFD-1-1 and CFD-1-32). We conjecture that this result is caused by the small number of classes in the data sets (i.e., SST2 has 2 classes, while AG-News 4 classes). Finally, the experimental findings on i.i.d. (see the illustrations in Figure~\ref{fig:distilbert_results_sst2_and_agnews_100} with $\alpha$=100.0) and non-i.i.d. (see the illustrations in Figure~\ref{fig:distilbert_results_sst2_and_agnews_100} with $\alpha$=1.0) data, show that CFD is robust to changes in the clients' data heterogeneity.

Next, we investigate the upstream and downstream communication cost (in MB) necessary to achieve a certain target accuracy across different levels of data heterogeneity (i.e., with $\alpha$=100.0 and 1.0). On the SST2 dataset, we set the target accuracy to 0.88, while for the AG-News dataset, we set it to 0.91. The experimental results are reported in Table~\ref{tab:distilbert_bits_to_target}. First, we observe that CFD, referred in Table~\ref{tab:distilbert_bits_to_target} as CFD-1-32 (i.e., 1-bit upstream and 32-bit downstream communication), effectively reduces the communication overhead with respect FA and FD. Second, the communication cost is slightly increased for CFD in two cases: (i) when soft-label encoding ($\text{CFD}_\Delta$) is applied to CFD, and (ii) with changes in clients' data heterogeneity (i.e., when $\alpha$ transitions from 100.0 to 1.0). Thereafter, we investigate a scenario of highest compression. That is, when data exchanged between the clients and the server is quantized to 1-bit (using Equation~(\ref{eq:onehot})) in both directions, upstream ($b_{up}=1$) and downstream ($b_{down}=1$) communication. The experimental findings are shown in Table~\ref{tab:distilbert_bits_to_target}, in the columns named CFD-1-1 and CFD$_{\Delta}$-1-1. From this data, observe that the upstream and downstream communication cost is similar for each case of clients' data heterogeneity (i.e., $\alpha$=100.0). However, communication overhead is noticeable by a small amount when the clients' data changes from i.i.d. ($\alpha$=100.0) to non-i.i.d. ($\alpha$=1.0), and soft-label encoding ($\text{CFD}_\Delta$) is applied to CFD. Notice that, when contrasting CFD-1-32 against CFD-1-1 in the context of upstream communication, in some cases the cost is higher for the latter (except for AG-News, with $\alpha$=1.0). Though, at first sight, these results contradict the intuition, they suggest that the FL optimization problem becomes harder when 1-bit quantization is applied in both communication directions (i.e., it takes more communication rounds to achieve the target accuracy for CFD-1-1 than CFD-1-32). Overall, these findings suggests that 1-bit upstream/downstream quantization does not affect the model performance up to a certain target accuracy, and thus, it represents a suitable technique in the context of FL, when the highest compression is required and communication bandwidth is limited. 

For additional results on the effect of the distillation data set size on the performance of CFD, we refer the reader to the supplementary materials (see Table~\ref{tab:distilbert_bits_to_target_50_20_10} and Figure~\ref{fig:distilbert_results_sst2_and_agnews_50_20_10}).

%
%


%
%


\section{Conclusion}
\label{sec7}

In this work we have explored the communication properties of Federated Distillation and shown that drastic compression gains are possible. For instance, on language modelling tasks, we demonstrated that our proposed Compressed Federated Distillation method can reduce the cumulative communication necessary to achieve fixed performance targets from 8570 MB to 0.81 MB when compared to the very popular Federated Averaging algorithm. This corresponds to a reduction in communication by $\times$ 10580. Similar compression rates were obtained in our investigated image classification problems on popular convolutional neural networks. We believe that our findings will help the widespread adoption of Federated Learning in heavily distributed and/or resource-constrained settings.

It is important to note however, that the favorable communication properties of all Federated Distillation methods, like the one considered in this paper, come at the cost of additional computational overhead caused by the local distillation. This additional computational overhead might be challenging in Federated Learning environments where client have limited computational resources, or where the number of clients is high and/or the number of data points per client is low. It thus needs to be carefully considered for every application, which of the two paradigms - Federated Averaging or Federated Distillation - is more suitable for the problem at hand.  

Federated Distillation is a very promising new way of solving Federated Learning problems, but many aspects of it are not fully understood yet. While it's unique communication properties and the added option for clients to train different local models could make it a popular choice for Federated Learning applications, it is also lacking formal robustness and convergence guarantees so far. Future work could address these open problems and also explore personalization techniques for FD via meta- or multi-task learning \cite{sattler2020clustered}.

\bibliographystyle{IEEEtran}
\bibliography{sample.bib}

\begin{IEEEbiography}[{\includegraphics[width=1in,height=1.25in,clip,keepaspectratio]{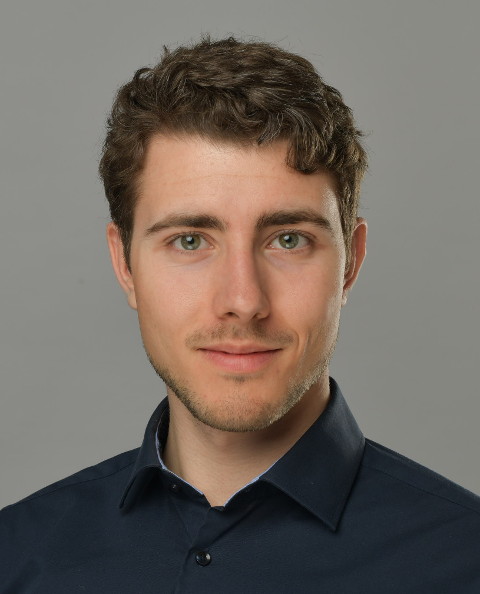}}]{Felix Sattler} received a M.Sc. degree in computer science, a M.Sc. degree in applied mathematics and a B.Sc. degree in Mathematics all from Technische Universit{\"a}t Berlin. He is currently with the Machine Learning Group, Fraunhofer Heinrich Hertz Institute, Berlin, Germany. His research interests include distributed machine learning, neural networks and multi-task learning.
\end{IEEEbiography}

\begin{IEEEbiography}[{\includegraphics[width=1in,height=1.25in,clip,keepaspectratio]{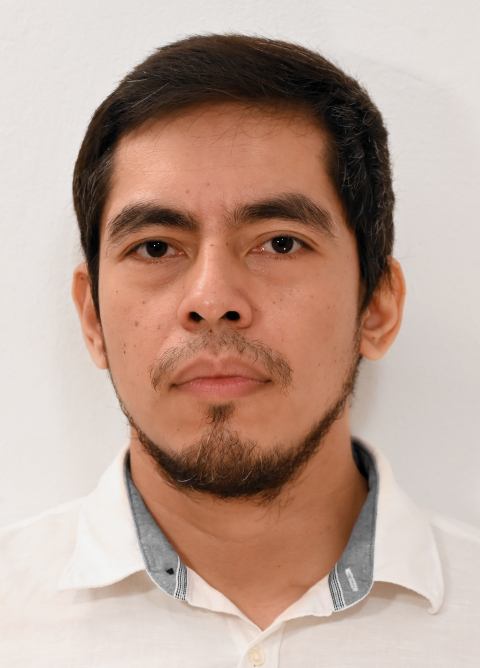}}]{Arturo Marban} received a B. Eng. degree in Mechatronics and M.Sc. degree in Manufacturing Systems from the Monterrey Institute of Technology and Higher Education, Monterrey, Mexico, in 2007 and 2010, respectively. In 2018, he received a Ph.D. degree in Automatic Control, Robotics, and Computer Vision from the Polytechnic University of Catalonia, Catalonia, Spain. Afterward, in the same year, he joined the Machine Learning Group, Fraunhofer Heinrich Hertz Institute, Berlin, Germany. His research interests include machine learning and neural networks, specifically, computer vision and efficient deep learning.
\end{IEEEbiography}

\begin{IEEEbiography}[{\includegraphics[width=1in,height=1.25in,clip,keepaspectratio]{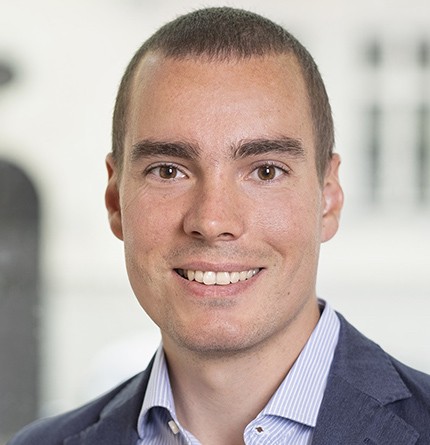}}]{Roman Rischke} received the M.Sc. degree in business mathematics from Technische Universität Berlin, Berlin, Germany, in 2012, and the Dr. rer. nat. degree in mathematics from Technische Universität München, Munich, Germany, in 2016. He currently works as a post-doctoral researcher in the Machine Learning Group at Fraunhofer Heinrich Hertz Institute, Berlin, Germany. His research interests include discrete optimization under data uncertainty, robust and trustworthy machine learning as well as distributed learning. 
\end{IEEEbiography}

\begin{IEEEbiography}[{\includegraphics[width=1in,height=1.25in,clip,keepaspectratio]{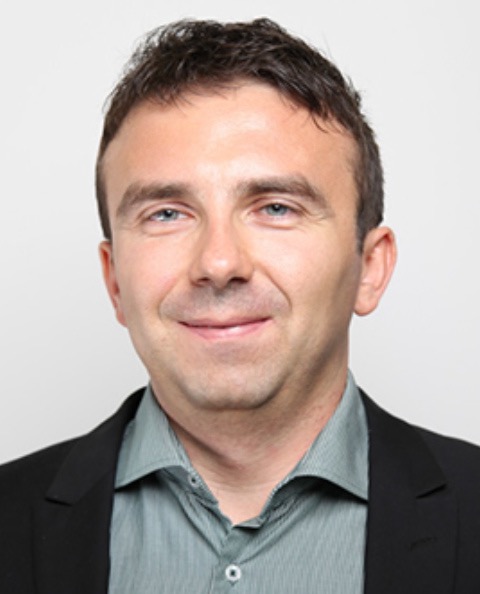}}]{Wojciech Samek} (M'13)
is head of the Machine Learning Group at Fraunhofer Heinrich Hertz Institute, Berlin, Germany. He received the Dipl.-Inf. degree in computer science from Humboldt University of Berlin, Germany, in 2010, and the Dr. rer. nat. degree from the Technical University of Berlin, Germany, in 2014. During his studies he was awarded scholarships from the Studienstiftung des deutschen Volkes and the DFG Research Training Group GRK 1589/1, and was a visiting researcher at NASA Ames Research Center, Mountain View, USA. He is PI at the Berlin Institute for the Foundation of Learning and Data (BIFOLD), member of the European Lab for Learning and Intelligent Systems (ELLIS) and associated faculty at the DFG graduate school BIOQIC. Furthermore, he is an editorial board member of Digital Signal Processing, PLoS ONE and IEEE TNNLS and an elected member of the IEEE MLSP Technical Committee. He is part of the MPEG-7 Part 17 standardization and was organizer of special sessions, workshops and tutorials on topics such as explainable AI and federated learning at top-tier machine learning and signal processing conferences. He has co-authored more than 100 peer-reviewed journal and conference papers, predominantly in the areas deep learning, explainable AI, neural network compression, and federated learning.\end{IEEEbiography}

\newpage

\section{Supplement}

\subsection{Additional Results}






Table~\ref{tab:distilbert_bits_to_target_50_20_10} describes the effect of the distillation dataset size in the server model performance, during upstream communication. Specifically, 50\%, 20\%, and 10\% of the distillation dataset samples are processed, and we report the upstream communication cost (in MB) necessary to achieve a certain target accuracy. For the SST2 dataset, the target accuracy was set to 0.88, while for the AG-News dataset, to 0.91. On the other hand, Figure~\ref{fig:distilbert_results_sst2_and_agnews_50_20_10} shows the complete communication cost dynamics (i.e., accuracy vs. communication cost at every communication round) for these experiments.

\input{lang_model_results/table_distilbert_50_20_10_bits_to_target}
\input{lang_model_results/fig_distilbert_results_sst2_and_agnews_50_20_10}

\end{document}

%% file: lang_model_results/table_distilbert_bits_to_target.tex
\begin{table*}[]
\caption{Upstream and downstream communication, measured in [MB], required in Federated Training of DistilBERT, to achieve a specific target accuracy, on the SST-2 and AG-News data sets at different levels of data heterogeneity $\alpha$.}
\centering
\begin{tabular}{llrrrrrrr}
\toprule
Dataset           & $\alpha$ & Up/Down  & FA            & FD        & CFD-1-32  & $\text{CFD}_\Delta$-1-32  & CFD-1-1   & $\text{CFD}_\Delta$-1-1 \\
(Target Accuracy) &          &          &               &           &           &                           &           &                            \\
\midrule
SST2              & 100.0    & Up       & 267.820 	& 0.270 & 0.004  & \textbf{0.005}  & 0.029 & 0.044 \\
(0.88)            & 100.0    & Down     & 267.820 	& 0.270 & 0.269  & 0.269 & \textbf{0.030} & 0.044  \\
                  & 1.0      & Up       & 803.460 	& 0.539 & \textbf{0.007}  & 0.011  & 0.031 & 0.046  \\
                  & 1.0      & Down     & 803.460 	& 0.539 & 0.539  & 0.539 & \textbf{0.034} & 0.050  \\
\midrule
AG-News           & 100.0    & Up       & 535.640 	& 1.920 & \textbf{0.030}  & 0.035  & 0.090 & 0.104  \\
(0.91)            & 100.0    & Down     & 535.640 	& 1.920 & 1.920 & 1.920 & \textbf{0.090} & 0.104  \\
                  & 1.0      & Up       & 1071.280 	& 4.800 & 0.142  & 0.169  & \textbf{0.101} & 0.119  \\
                  & 1.0      & Down     & 1071.280 	& 4.800 & 9.600 & 9.600 & \textbf{0.105} & 0.121  \\
\bottomrule
\end{tabular}
\label{tab:distilbert_bits_to_target}
\end{table*}

%% file: lang_model_results/fig_distilbert_results_sst2_and_agnews_100.tex
\begin{figure}
    \centering
    \begin{subfigure}[b]{0.485\textwidth}
        \centering
        \includegraphics[width=1.\linewidth]{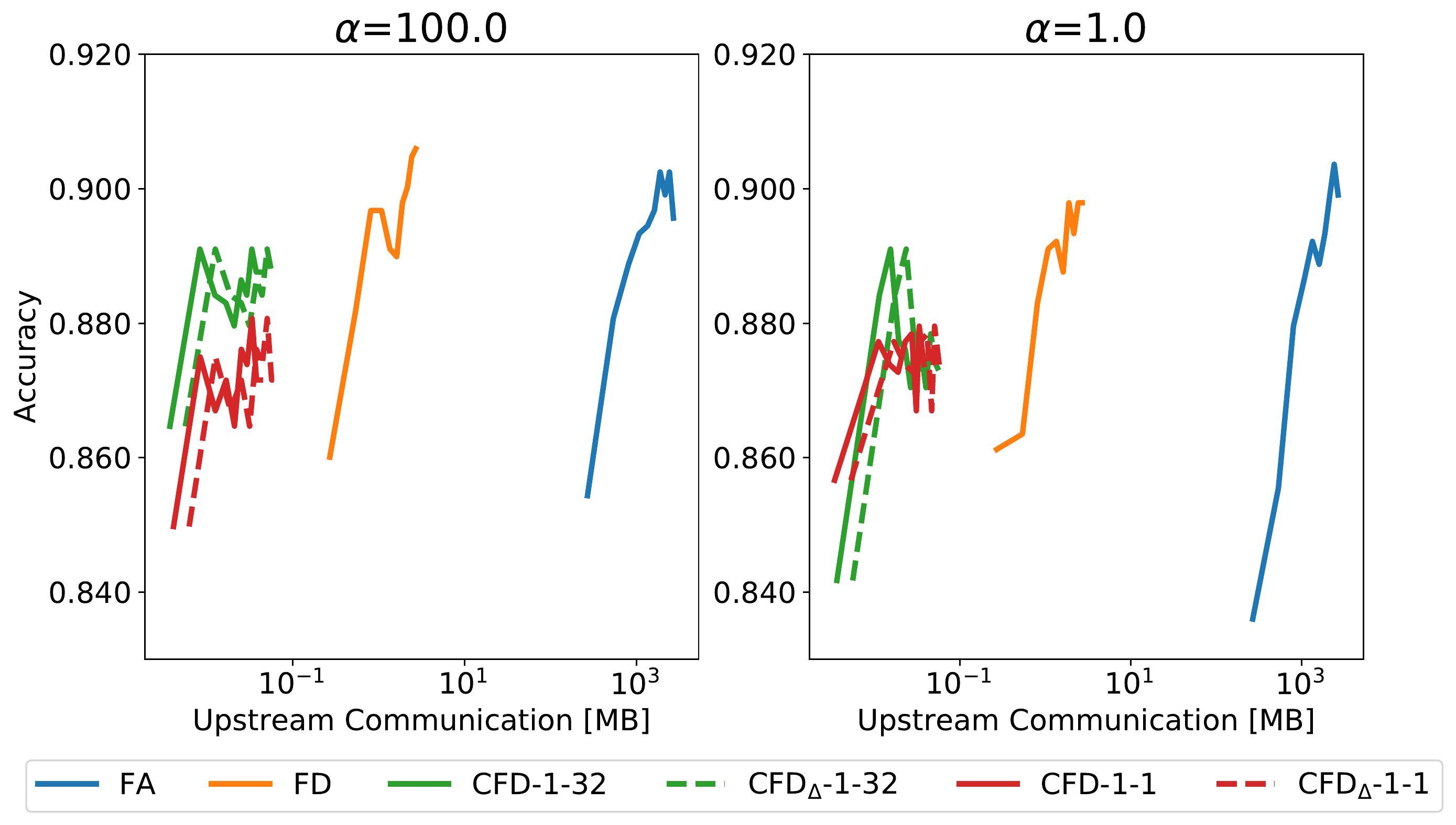}
        \caption{SST2.}
        \label{fig:distilbert_results_sst2_100}
    \end{subfigure}
    \begin{subfigure}[b]{0.485\textwidth}
        \centering
        \includegraphics[width=1.\linewidth]{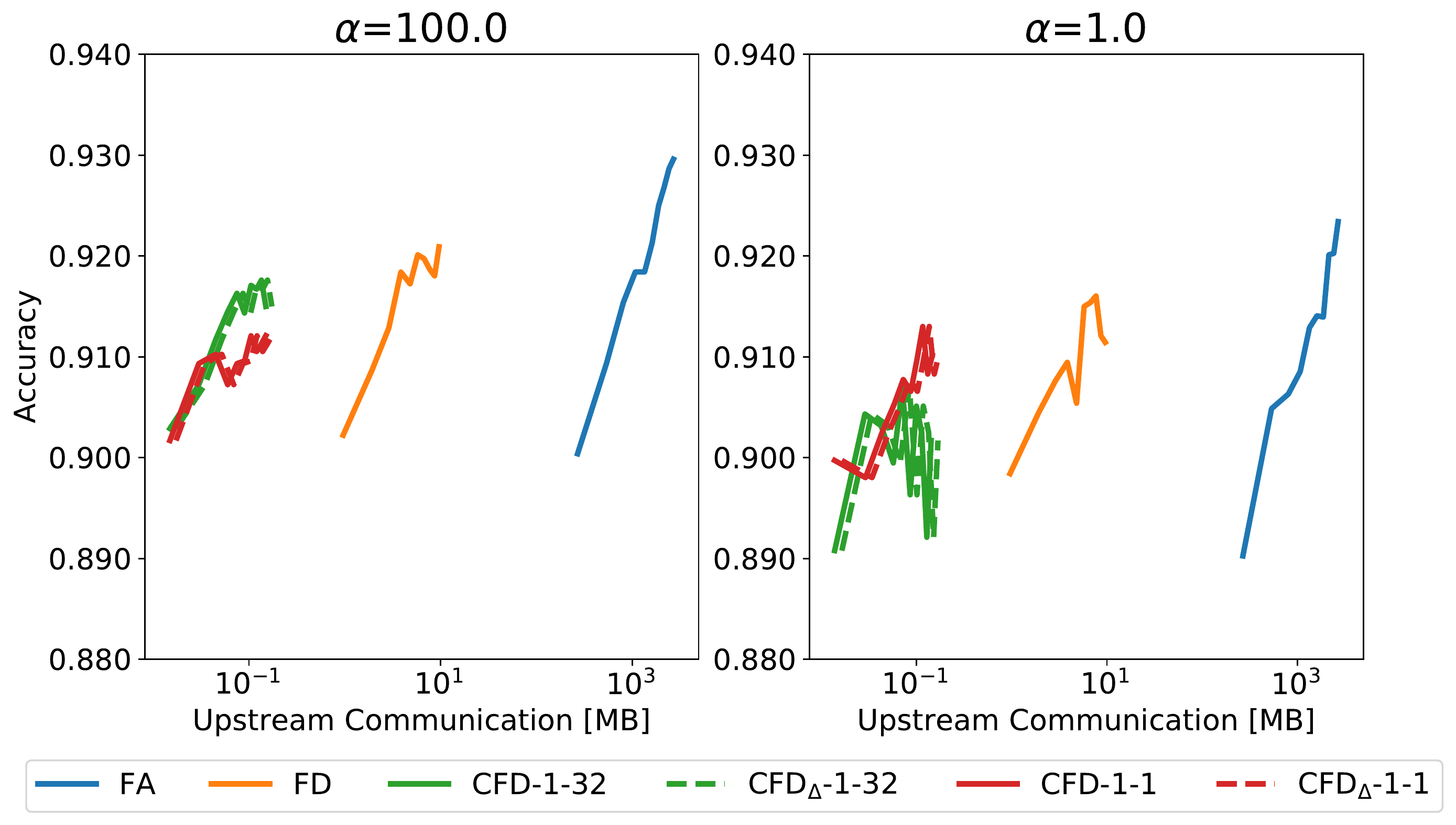}
        \caption{AG-News.}
        \label{fig:distilbert_results_agnews_100}
    \end{subfigure}
    \caption{DistilBERT model performance as a function of communicated bits for our proposed CFD method and baselines FA and FD. This model was evaluated on the SST2 and AG-News datasets, with $\alpha$ = 100.0 and 1.0, using 100\% of the disitillation dataset samples (10 clients with 100\% participation rate).}
    \label{fig:distilbert_results_sst2_and_agnews_100}
\end{figure}

%% file: lang_model_results/table_distilbert_50_20_10_bits_to_target.tex
\begin{table*}[!ht]
\caption{Upstream communication measured in [MB], necessary to achieve a specific target accuracy in Federated Learning of DistillBERT, on the SST2 and AG-News data set, at different numbers of distillation data samples (50\%, 20\%, and 10\%), and levels of data heterogeneity ($\alpha$=100.0 and 1.0).}
\centering
\begin{tabular}{llrrrrr}
\toprule
Dataset             & Distillation  & $\alpha$  & FA            & FD        & CFD-1-32       & $\text{CFD}_\Delta$-1-32 \\
(Target Accuracy)   & Dataset Size  &           &               &           &           & \\
\midrule
SST2    & 50\%      & 100.0 & 267.820  & 0.135 & 0.003 & 0.003 \\
(0.88)  &           & 1.0   & 803.460  & 0.404 & 0.004 & 0.005 \\
        & 20\%      & 100.0 & 267.820  & 0.161 & 0.003 & 0.004 \\
        &           & 1.0   & 803.460  & 0.161 & 0.003 & 0.005 \\
        & 10\%      & 100.0 & 267.820  & 0.135 & 0.001 & 0.001 \\
        &           & 1.0   & 803.460  & 0.161 & 0.004 & 0.006 \\
\midrule
AG-News & 50\%      & 100.0 & 535.640  & 0.480 & 0.015 & 0.018 	\\
(0.91)  &           & 1.0   & 1071.280 & 1.920 & 0.071 & 0.084 	\\
        & 20\%      & 100.0 & 535.640  & 0.384 & 0.003 & 0.004 	\\
        &           & 1.0   & 1071.280 & 1.536 & 0.029 & 0.034 	\\
        & 10\%      & 100.0 & 535.640  & 0.384 & 0.007 & 0.009 	\\
        &           & 1.0   & 1071.280 & 0.960 & 0.014 & 0.018 	\\
\bottomrule
\end{tabular}
\label{tab:distilbert_bits_to_target_50_20_10}
\end{table*}

%% file: lang_model_results/fig_distilbert_results_sst2_and_agnews_50_20_10.tex
\begin{figure*}[!ht]
    \centering
    \begin{subfigure}[b]{0.485\textwidth}
        \centering
        \includegraphics[width=1.\linewidth]{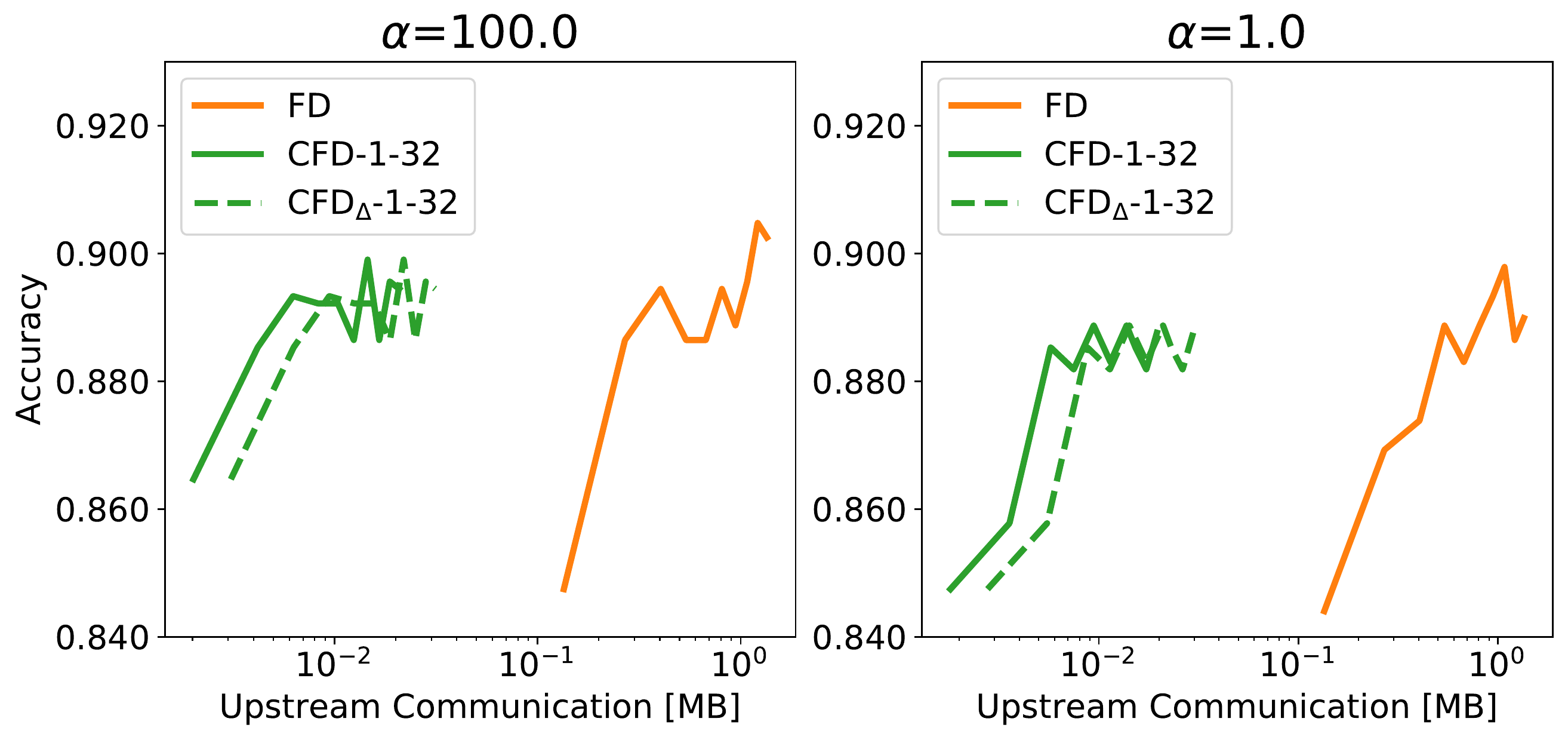}
        \caption{SST2 (50\% distillation data samples).}
        \label{fig:distilbert_results_sst2_50}
    \end{subfigure}
    \hfill
    \begin{subfigure}[b]{0.485\textwidth}
        \centering
        \includegraphics[width=1.\linewidth]{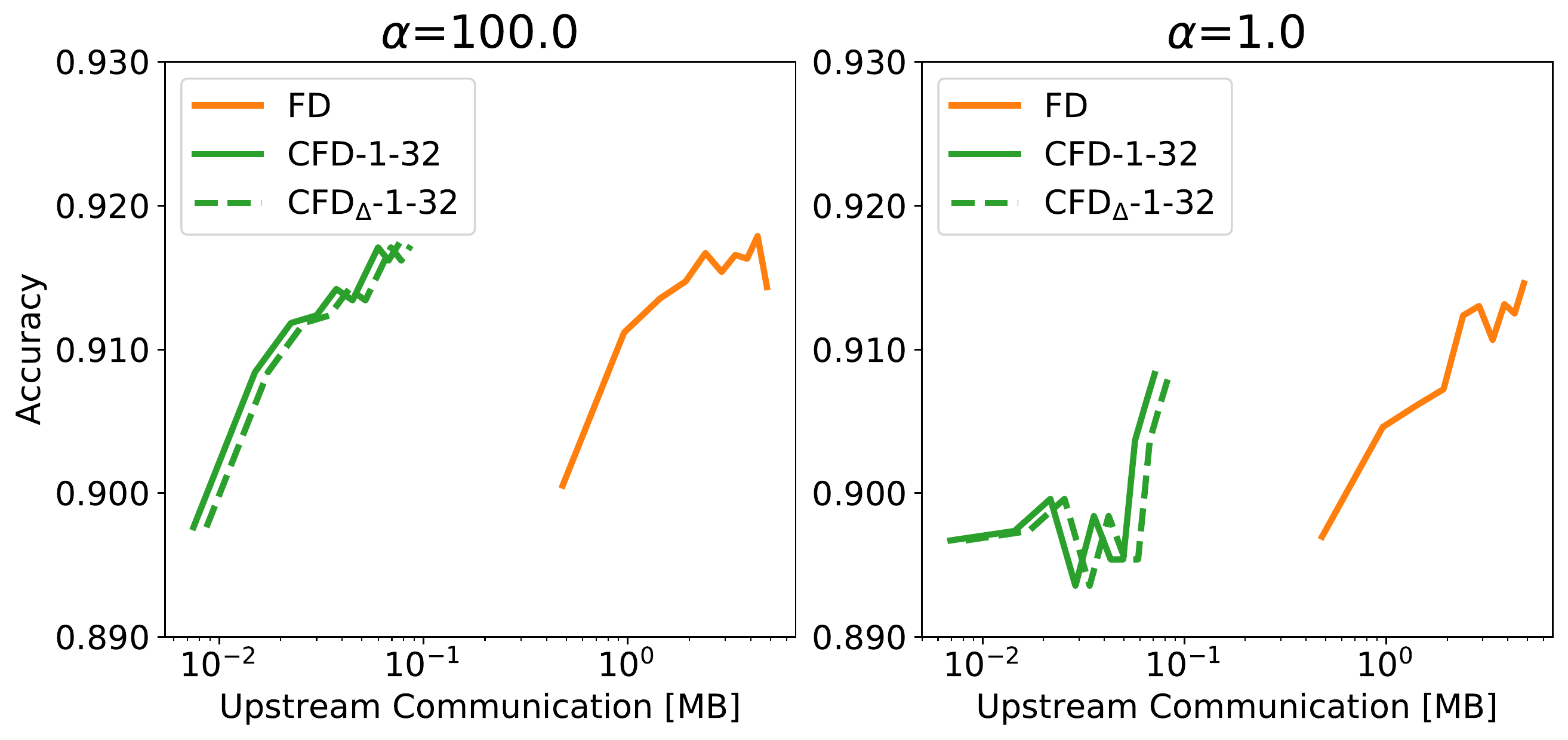}
        \caption{AG-News (50\% distillation data samples).}
        \label{fig:distilbert_results_agnews_50}
    \end{subfigure} 
    \newline~\newline
    \begin{subfigure}[b]{0.485\textwidth}
        \centering
        \includegraphics[width=1.\linewidth]{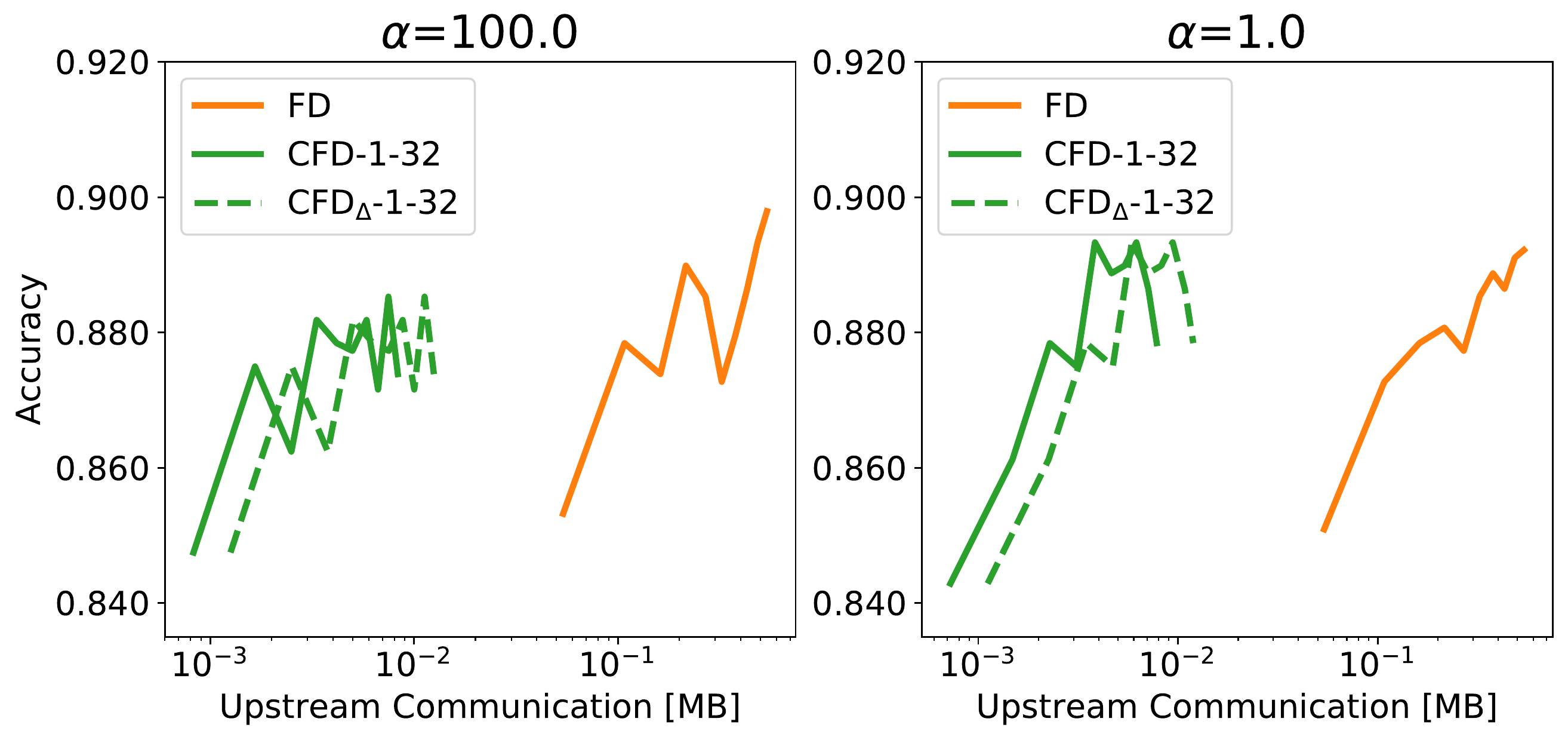}
        \caption{SST2 (20\% distillation data samples).}
        \label{fig:distilbert_results_sst2_20}
    \end{subfigure}
    \hfill
    \begin{subfigure}[b]{0.485\textwidth}
        \centering
        \includegraphics[width=1.\linewidth]{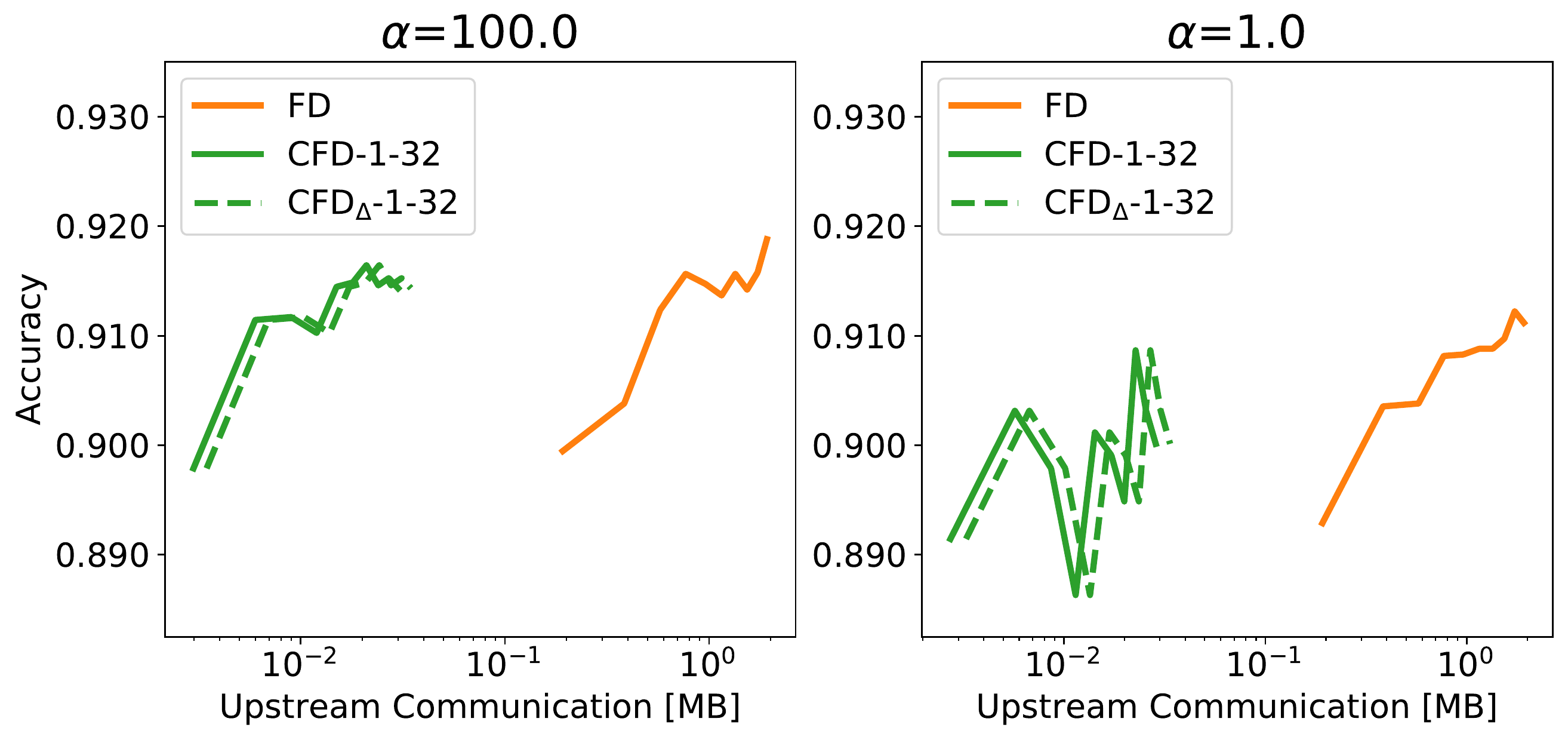}
        \caption{AG-News (20\% distillation data samples).}
        \label{fig:distilbert_results_agnews_20}
    \end{subfigure}
    \newline~\newline
    \begin{subfigure}[b]{0.485\textwidth}
        \centering
        \includegraphics[width=1.\linewidth]{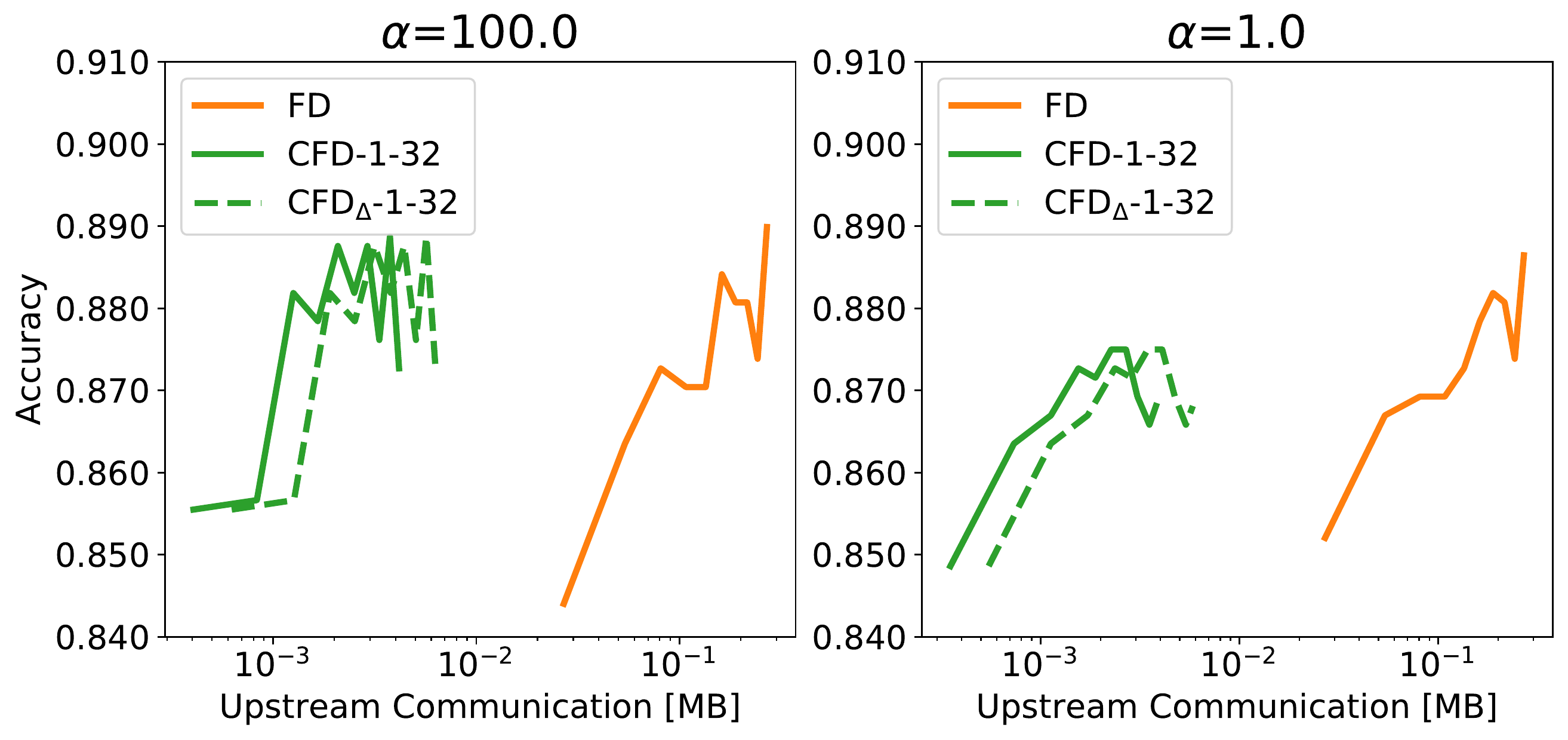}
        \caption{ SST2 (10\% distillation data samples). }
        \label{fig:distilbert_results_sst2_10}
    \end{subfigure}
    \hfill
    \begin{subfigure}[b]{0.485\textwidth}
        \centering
        \includegraphics[width=1.\linewidth]{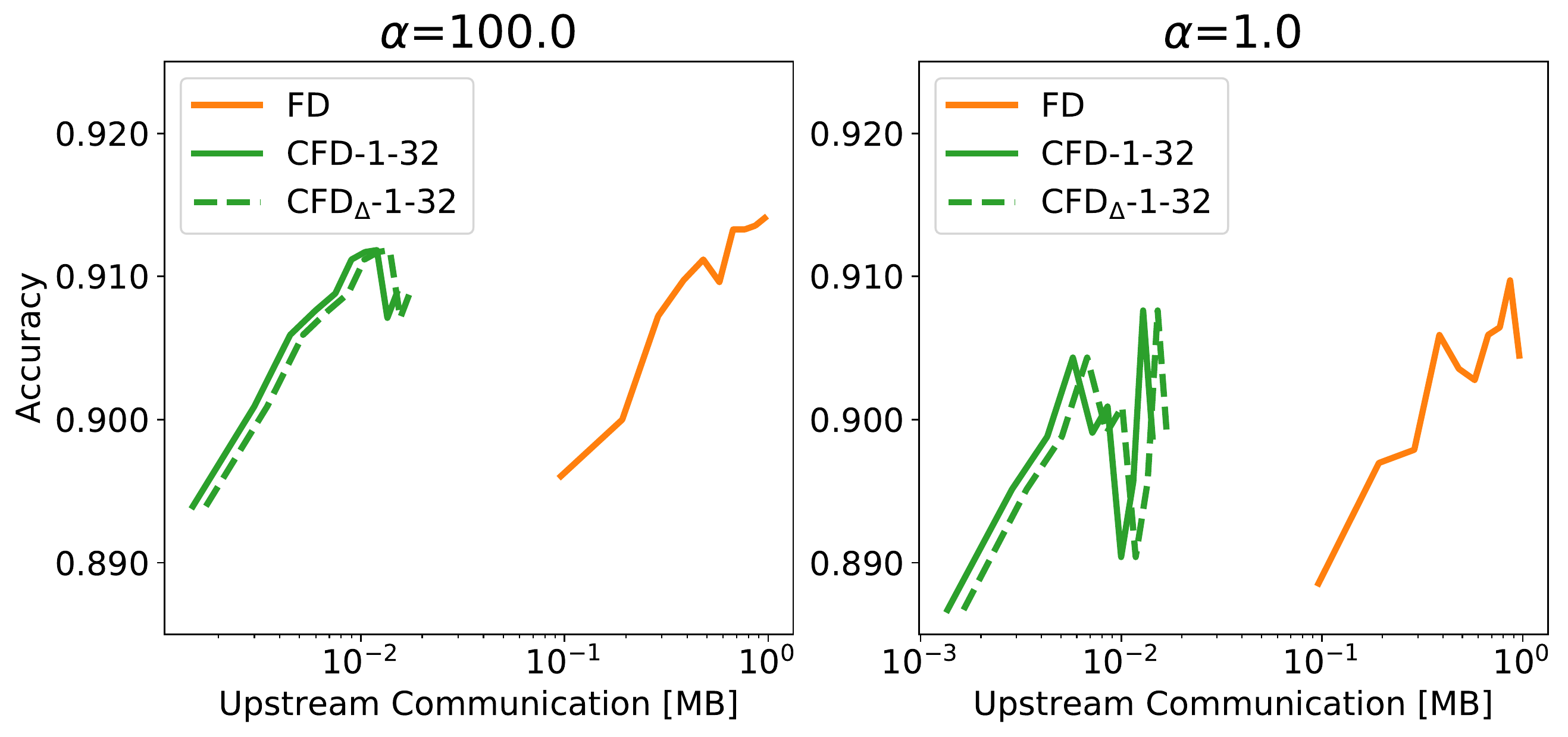}
        \caption{AG-News (10\% distillation data samples).}
        \label{fig:distilbert_results_agnews_10}
    \end{subfigure}
    \caption{Communication efficiency (i.e., accuracy vs communication cost) for Federated Learning of DistilBERT, on the SST2 and AG-News datasets, with $\alpha$ = 100.0 and 1.0, using 50\%, 20\%, and 10\% of the disitillation dataset samples (10 clients with 100\% participation rate).}
    \label{fig:distilbert_results_sst2_and_agnews_50_20_10}
\end{figure*}